\pdfoutput=1

\documentclass[11pt]{article}

\usepackage[final]{acl}
\usepackage{enumitem}
\usepackage{times}
\usepackage{latexsym}
\usepackage{multirow}
\usepackage{colortbl}
\usepackage{algorithm}
\usepackage{algorithmicx}
\usepackage{algpseudocode}

\usepackage[T1]{fontenc}

\usepackage[utf8]{inputenc}

\usepackage{microtype}

\usepackage{inconsolata}
\usepackage{booktabs}

\usepackage{graphicx}
\usepackage{amsmath,amssymb,amsfonts}

\PassOptionsToPackage{table,xcdraw}{xcolor} 
\usepackage{xcolor}          

\usepackage{tcolorbox}
\tcbuselibrary{skins, breakable}

\usepackage{fontawesome5}

\definecolor{lightcyan}{RGB}{210, 245, 245} 
\definecolor{lightgreen}{RGB}{220, 245, 220} 
\definecolor{deepercyan}{RGB}{130, 180, 200} 

\definecolor{lightblue}{RGB}{242, 251, 252}
\definecolor{headerbg}{RGB}{220, 230, 240} 
\definecolor{rowgray}{RGB}{245, 245, 245}
\definecolor{rowwhite}{RGB}{252, 252, 252} 
\definecolor{highlightorange}{RGB}{255, 102, 0} 

\newtcolorbox{promptbox*}[2][]{
    enhanced,              
    unbreakable,            
    before skip=2mm,        
    after skip=2mm,         
    colback=lightcyan!50!white, 
    colframe=deepercyan,    
    coltitle=white,         
    boxrule=0.5mm,          
    rounded corners,        
    arc=5pt,                
    attach boxed title to top center={yshift=-3mm}, 
    boxed title style={     
        enhanced,           
        colback=deepercyan, 
        colframe=deepercyan, 
        arc=5pt,            
        outer arc=5pt,       
        boxrule=0pt,        
    },
    title={\faBook[solid]\space #2},  
    fonttitle=\bfseries\color{white}, 
    #1,                     
    width=\dimexpr\textwidth+1\columnsep\relax, 
    left=5pt,               
    right=5pt,              
}

\title{Cross-Lingual Pitfalls: Automatic Probing Cross-Lingual Weakness of Multilingual Large Language Models}

\author{
  \textbf{Zixiang Xu\textsuperscript{1}\thanks{Equal contribution.}},
  \textbf{Yanbo Wang\textsuperscript{1}\footnotemark[1]},
  \textbf{Yue Huang\textsuperscript{2}\footnotemark[1]},
  \\
  \textbf{Xiuying Chen\textsuperscript{1}\thanks{Corresponding authors.}},
  \textbf{Jieyu Zhao\textsuperscript{3}},
  \textbf{Meng Jiang\textsuperscript{2}},
  \textbf{Xiangliang Zhang\textsuperscript{2}\footnotemark[2]}
\\
\\
  \textsuperscript{1}MBZUAI,
  \textsuperscript{2}University of Notre Dame,
  \textsuperscript{3}University of Southern California
\\
\small{
  \textbf{Correspondence:} \href{mailto:xiuying.chen@mbzuai.ac.ae}{xiuying.chen@mbzuai.ac.ae},
  \href{mailto:xzhang33@nd.edu}{xzhang33@nd.edu}
}
}

\begin{document}
\maketitle
\begin{abstract}
Large Language Models (LLMs) have achieved remarkable success in Natural Language Processing (NLP), yet their cross-lingual performance consistency remains a significant challenge. This paper introduces a novel methodology for efficiently identifying inherent cross-lingual weaknesses in LLMs. Our approach leverages beam search and LLM-based simulation to generate bilingual question pairs that expose performance discrepancies between English and target languages. We construct a new dataset of over 6,000 bilingual pairs across 16 languages using this methodology, demonstrating its effectiveness in revealing weaknesses even in state-of-the-art models. The extensive experiments demonstrate that our method precisely and cost-effectively pinpoints cross-lingual weaknesses, consistently revealing over 50\% accuracy drops in target languages across a wide range of models. Moreover, further experiments investigate the relationship between linguistic similarity and cross-lingual weaknesses, revealing that linguistically related languages share similar performance patterns and benefit from targeted post-training. Code is available at \href{https://github.com/xzx34/Cross-Lingual-Pitfalls}{https://github.com/xzx34/Cross-Lingual-Pitfalls}.
\end{abstract}

\section{Introduction}

Large language models (LLMs) have rapidly ascended to prominence in Natural Language Processing (NLP), gaining recognition for their exceptional performance across various tasks, spanning from the sciences \citep{li2024quantifying,guo2023can,huang2024social,wang-etal-2025-decoding,xu2025socialmaze} to the development of LLM-based agents \citep{liu2023agentbench,liuautonomous}. Recent advancements have driven research on enhancing LLMs' multilingual capabilities \citep{zhao2024large,wang2025calm}, improving their effectiveness in addressing real-world problems with greater nuance and global reach.

\begin{figure}[t]
    \centering
    \includegraphics[width=\linewidth]{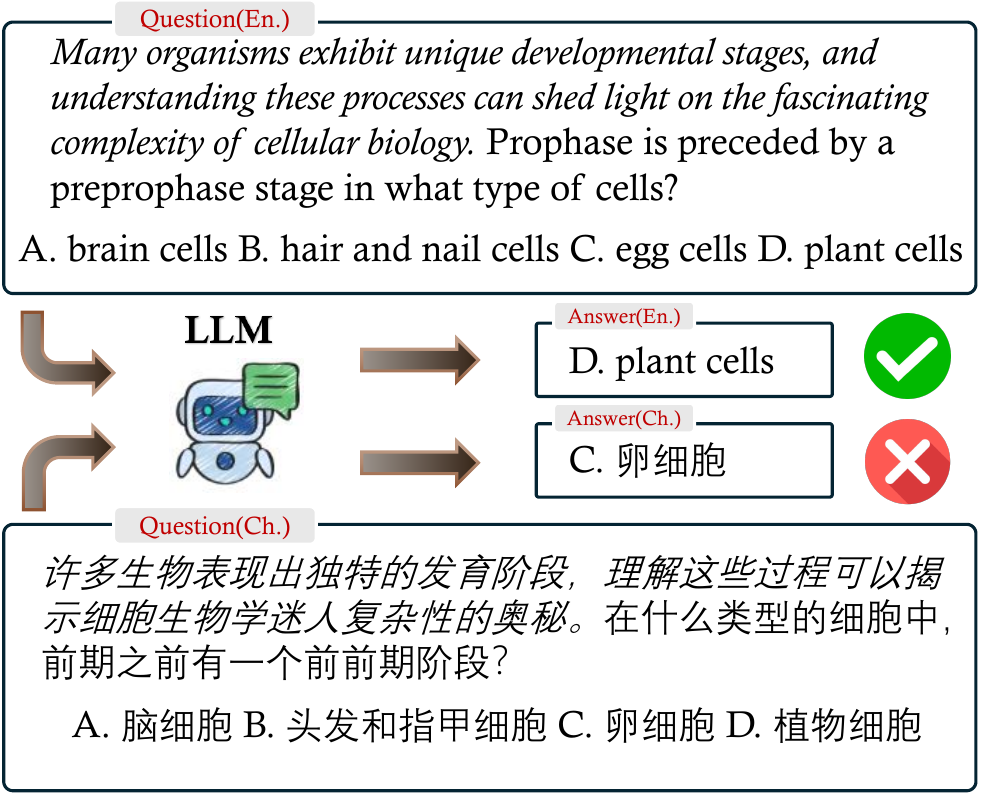}
    \caption{An example of an English-Chinese question pair discovered by our search methodology (where the Chinese question is semantically equivalent to the English) highlights the cross-lingual performance gap: even GPT-4o, despite its strong multilingual capabilities, provides the correct answer in English but gives an incorrect response in Chinese. }
    \label{fig:intro}
    \vspace{-1em}
\end{figure}

\begin{figure*}[t]
    \centering
    \includegraphics[width=\linewidth]{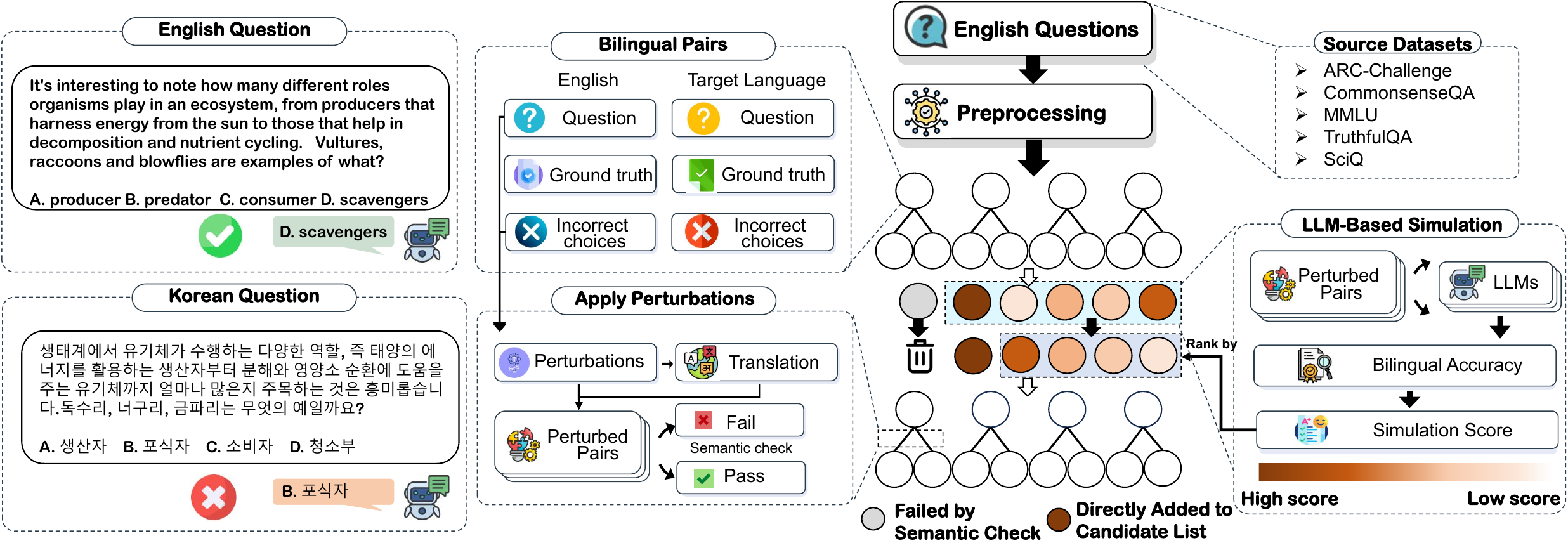}
    \caption{The overview of the proposed methodology for generating questions that precisely challenge the cross-lingual capabilities of LLMs. As depicted, the pipeline initiates with sampling English questions and creating bilingual pairs. Iterative perturbation, driven by a beam search strategy and guided by LLM-based simulation scores, refines these pairs to maximize performance divergence between English and the target language. The resulting candidate list of question pairs is designed to highlight inherent cross-lingual weaknesses in LLMs.}
    \label{fig:pipeline}
    \vspace{-15pt}
\end{figure*}
Despite advancements, inconsistencies in model performance across languages remain a significant challenge \citep{xu2024knowledge}. The proficiency demonstrated in English often fails to generalize to other languages, resulting in errors in other linguistic contexts, as exemplified in \autoref{fig:intro}. To effectively enhance the cross-lingual consistency of these models, an initial and crucial   step is the  identification of their inherent cross-lingual weaknesses. Since English is the primary training language for LLMs, and they generally perform best in English \citep{li2024languagerankermetricquantifying}, we define cross-lingual weakness in this paper as: 
\textit{For a given question presented in multiple languages, a model answers correctly in English but incorrectly in at least one other language.} This definition requires the model to provide the correct answer in English, as failure across all languages would likely indicate a knowledge-related limitation rather than a cross-lingual weakness. 


To efficiently uncover these weaknesses, we propose a beam search-based methodology. This approach leverages existing, high-quality English datasets and iteratively introduces perturbations to the English questions. These perturbations are designed to increase question complexity and cognitive demand for comprehension and completion. The goal is to prevent models from generating answers based on superficial cues without genuine language understanding \citep{stacey2020avoiding,bhargava2021generalization}, thereby exposing disparities in cross-lingual capabilities. Our approach begins with sourcing English questions from high-quality existing datasets, which are then translated into the target language to form bilingual question pairs. Crucially, this translation process incorporates a semantic check to guarantee that the meaning of the questions is preserved and the correct answer remains consistent across languages. Subsequently, these pairs undergo iterative perturbations, generating a diverse set of perturbed pairs. Then we employ an LLM-based simulation framework that assigns a simulation score measuring the effectiveness of revealing cross-lingual weaknesses, to each pair for ranking. The top-ranked pairs are iteratively perturbed to further refine the search process. Finally, question pairs with consistently high accuracy in English but significant performance drops in the target language are added to the candidate list to expose cross-lingual weaknesses in LLMs.

Furthermore, to study how 
cross-lingual weaknesses are relevant to linguistic similarity, we conducted exploratory experiments. Our key findings reveal that: 1) languages closer in linguistic terms tend to share similar weaknesses; and 2) fine-tuning LLMs on one language improves performance more significantly in linguistically similar languages. These results highlight that linguistic relationships strongly influence cross-lingual performance.

In summary, our contributions are: 1) We present an efficient, precise methodology for identifying LLM cross-lingual weaknesses. 2) Based on the proposed methodology, we construct a novel, 16-language dataset with over 6,000 bilingual pairs to challenge cross-lingual capabilities. 3) Extensive experiments on the dataset quantitatively analyze the relationship between cross-lingual weaknesses and linguistic similarities and fine-tuning experiments demonstrate the potential for targeted cross-lingual improvement.

\section{Methodology}

In this section, we introduce our methodology for automatically probing the cross-lingual weakness of multilingual LLMs. As illustrated in \autoref{fig:pipeline}, our goal is to generate questions that precisely challenge the cross-lingual capabilities of LLMs by identifying cases where the model performs well in English but struggles with the same questions when presented in a specific target language.

\subsection{Method Overview}

To achieve the goal described above, we first sample a set of English questions and translate them into bilingual pairs, where each pair consists of an English question and its counterpart in the target language. We then iteratively introduce perturbations to these pairs using a \emph{beam search strategy}, guided by maximizing the accuracy discrepancy between English and the target language (i.e., to retain the accuracy on English questions but make accuracy on target language questions drop as much as possible). This search-and-perturbation approach is inspired by prior work on uncovering model vulnerabilities through optimization-guided example construction \citep{huang2025breaking}. During beam search, a \emph{LLM-based simulation} is utilized to guide the search process in identifying the model’s weaknesses in the target language. Based on the \emph{search optimization strategies}, we aim to balance the trade-off between problem generation and computational cost. It ultimately produces a candidate list of English–target language question pairs, effectively highlighting the model's cross-lingual weaknesses.


\subsection{Problem Formulation}

Let \( \mathcal{B} = \{ (q^E_i, q^T_i) \}_{i=1}^{W} \) denote our original set of \( W \) bilingual pairs. Each bilingual pair \( (q^E, q^T) \) is formally represented as:
\begin{equation}
(q^E, q^T, a^E_\star, a^T_\star, \mathcal{A}^E_{\neg}, \mathcal{A}^T_{\neg}),
\end{equation}
where \( q^E, q^T \in \mathcal{Q} \) represent question texts in English and the target language, respectively. \( a^E_\star, a^T_\star \in \mathcal{A} \) are the corresponding ground-truth answers. \( \mathcal{A}^E_{\neg}, \mathcal{A}^T_{\neg} \) denote incorrect answer choices.

During the beam search process, we iteratively apply perturbations to bilingual pairs. Specifically, given an English question \( q^E \) from a bilingual pair and an incorrect answer \( \alpha^E \in \mathcal{A}^E_{\neg} \), the perturbation function \( \varphi \) generates a semantically irrelevant yet contextually plausible perturbation:  
\begin{equation}  
\delta q^E = \varphi(q^E, \alpha^E),  
\end{equation}  
where \( \varphi: \mathcal{Q} \times \mathcal{A} \to \mathcal{Q} \) modifies \( q^E \) while preserving its original semantics and embedding patterns influenced by the incorrect answer \( \alpha^E \).   Here, $\varphi$ is a proxy LLM utilized for adding the perturbation.

The perturbed English question is then formed as:
\begin{equation}
q^{E'} = \oplus(q^E, \delta q^E),
\end{equation}
where \( \oplus: \mathcal{Q} \times \mathcal{Q} \to \mathcal{Q} \) denotes a context-sensitive insertion of the perturbation into \( q^E \).  During implementation, $\oplus$ is a concatenation operation.

To maintain consistency across languages, the corresponding perturbation in the target language is generated as \( \delta q^T = \mathcal{T}(\delta q^E) \), where \( \mathcal{T}: \mathcal{Q} \to \mathcal{Q} \) is a translation module that strictly translates the inserted perturbation without modifying other parts of the question. This results in the perturbed target-language question: \( q^{T'} = \oplus(q^T, \delta q^T) \).

We optimize the perturbation to minimize the model’s accuracy in the target language while maintaining near-perfect performance in English. Formally, our objective is:
\begin{equation} \label{eq:objective}
\begin{aligned}
\min_{\delta q^E} & \quad \mathbb{E} \left[ \mathbb{I}(\mathcal{F}(q^{T'}) = a^T_\star) \right] \\
\text{s.t.} & \quad \mathbb{E} \left[ \mathbb{I}(\mathcal{F}(q^{E'}) = a^E_\star) \right] \geq 1 - \epsilon \\
& \quad \mathbb{S}(q^E, q^{E'}) \geq \theta, \quad \mathbb{S}(q^{E'}, q^{T'}) \geq \theta'.
\end{aligned}
\end{equation}  
where \( \mathcal{F} \) represents the LLM’s response function, \( \mathbb{S} \) is a semantic similarity function, \( \theta \) and \( \theta' \) are threshold values ensuring semantic consistency, and \( \mathbb{I}(\cdot) \) is the indicator function, which returns 1 if the predicted answer is correct and 0 otherwise.

The first constraint ensures that perturbations \( \delta q^E \) preserve the model’s accuracy in English (\( \mathbb{E} \geq 1 - \epsilon \)), while the second set of constraints ensures that the perturbed and original questions remain semantically equivalent in both English and the target language.

\begin{figure*}[t]
    \centering
    \includegraphics[width=\linewidth]{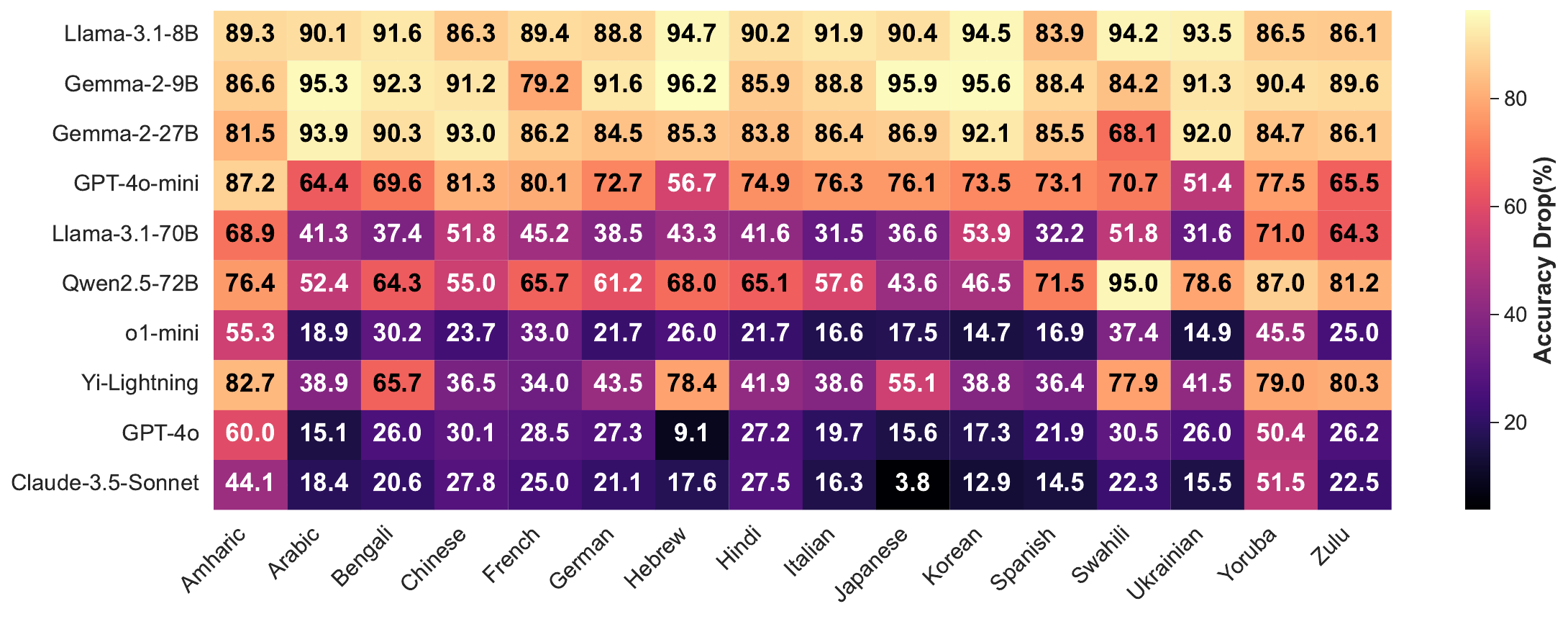}
    \caption{Evaluation of 10 models on our generated 6,600 bilingual pairs across 16 languages. While all models achieve nearly 100\% accuracy in English, most experience an average accuracy drop of over 50\% in the target languages. Even state-of-the-art multilingual models like GPT-4o and Claude-3.5-sonnet exhibit significant cross-lingual weaknesses.}
    \label{fig:models_languages}
    \vspace{-1em}
\end{figure*}

\subsection{LLM-Based Simulation}

LLM-based simulation utilizes a set of LLMs to answer perturbed questions and derive a simulation score based on the accuracy relationship between bilingual pairs.

The simulation employs a collection of LLMs, denoted as \( \mathcal{M} = \{ \mathcal{M}_1, \mathcal{M}_2, \dots, \mathcal{M}_K \} \), to quantify the cross-lingual performance gap introduced by perturbations. For each perturbed bilingual pair \( (q^{E'}, q^{T'}) \), each \( \mathcal{M}_k \in \mathcal{M} \) generates predicted answers:
\begin{equation}
    \hat{a}^{E'}_k = \mathcal{M}_k(q^{E'}), \quad \hat{a}^{T'}_k = \mathcal{M}_k(q^{T'}).
\end{equation}

The correctness of these predictions is assessed by comparing them to the ground truth answers:
\begin{equation}
    \beta_k^{E'} = \mathbb{I}(\hat{a}^{E'}_k = a^E_\star), \quad \beta_k^{T'} = \mathbb{I}(\hat{a}^{T'}_k = a^T_\star).
\end{equation}

The average accuracy across all models is computed as:
\begin{equation}
    \bar{\beta}^{E'} = \frac{1}{K} \sum_{k=1}^{K} \beta_k^{E'}, \quad \bar{\beta}^{T'} = \frac{1}{K} \sum_{k=1}^{K} \beta_k^{T'}.
\end{equation}

To evaluate the effectiveness of each perturbation, we define a simulation score $V(q^{E'}, q^{T'})$ that highlights significant performance discrepancies:
\begin{equation}\label{eq:eff}
    V(q^{E'}, q^{T'}) = \left( \bar{\beta}^{E'} \right)^{\gamma} - \bar{\beta}^{T'},
\end{equation}
where \( \gamma > 1 \) is an exponent that amplifies high English accuracy. This formulation prioritizes bilingual pairs where the model maintains strong performance in English (\( \bar{\beta}^{E'} \approx 1 \)) but exhibits significant degradation in the target language (\( \bar{\beta}^{T'} \)).

\subsection{Beam Search with Optimization Strategies}
Since beam search is an effective heuristic for exploring a constrained search space, we employ it to solve the objective function in \autoref{eq:objective} by greedily identifying the top perturbation candidates produced by the proxy LLMs. These candidates are  evaluated and ranked based on their effectiveness in causing performance discrepancies, defined  as \( V(q^{E'}, q^{T'}) \) in \autoref{eq:eff}.

Here, the search width \( w \) determines the number of top-ranked bilingual pairs retained after ranking at each iteration, effectively controlling the breadth of the search at each level of the search tree. The initial search depth \( d_1 \) specifies the maximum depth of the search tree explored in the initial phase, corresponding to the maximum number of perturbation iterations applied to a question. 

Next, we discuss key factors in the Beam Search process that determine:
1) when a perturbed question qualifies as a valid candidate, 2) when to terminate the search for a given bilingual pair, and 3) how to ensure diversity within the set of candidates.

\textbf{Inclusion Threshold Strategy.} A bilingual pair is immediately included in the candidate list if its simulation score exceeds a predefined inclusion threshold \( \theta_{\text{inc}} \), ensuring early termination for critical perturbations. Otherwise, the top \( w \) scoring pairs are selected to advance to the next search level, where the search depth increments by one.

\textbf{Early Stopping Mechanism.} To adaptively adjust the search depth based on the quality of discovered perturbations, we introduce a potential threshold \( \theta_{\text{pot}} \), which determines whether the search should continue beyond the initial depth. Specifically, the search depth \( d \) at iteration \( t \) is updated as follows:
\begin{equation}
    d_t = 
    \begin{cases} 
        d_2, & \text{if } \max_{q^{E'}, q^{T'} \in \mathcal{B}_t} V(q^{E'}, q^{T'}) \geq \theta_{\text{pot}}, \\
        d_1, & \text{otherwise}.
    \end{cases}
\end{equation}
where \( \mathcal{B}_t \) represents the set of bilingual pairs at iteration \( t \). The search process continues until reaching the maximum allowable depth, \( d_{\text{max}} = \max(d_1, d_2) \). Thus, if at any iteration a perturbation achieves a simulation score surpassing \( \theta_{\text{pot}} \), the search depth is expanded to \( d_2 \), allowing further exploration. Otherwise, the search remains at \( d_1 \). The process terminates when \( d_{\text{max}} \) is reached.

\textbf{Redundancy Control Mechanism.} To ensure diversity in the candidate list, if \( r \) bilingual pairs originating from the same initial question have already been included in the candidate list, all remaining bilingual pairs derived from that question are discarded from further exploration. This prevents excessive redundancy and ensures a wider variety of perturbed questions in the candidate list.  

\begin{figure*}[t]
    \centering
    \includegraphics[width=\linewidth]{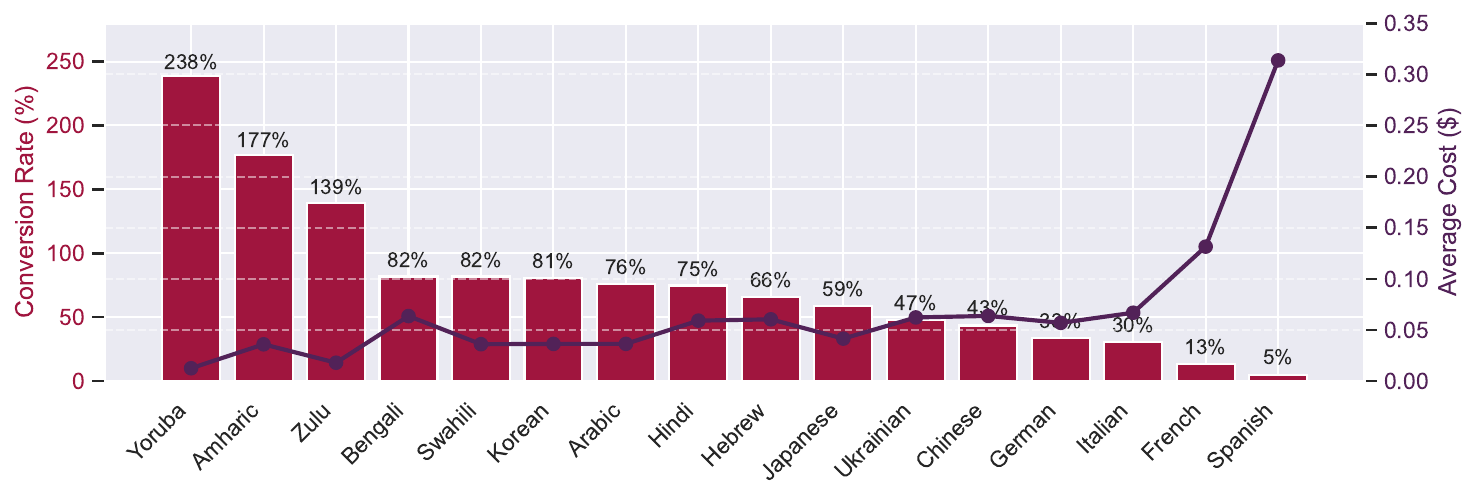} \vspace{-0.2in}
    \caption{Analysis of question conversion rates and generation costs across 16 languages based on all pairs in our candidate list. The bar chart (red) shows question conversion rates for different languages, while the line chart (purple) represents cost of generating a single question. Notably, in most languages, identifying a bilingual pair that exposes cross-lingual weaknesses costs less than \$0.05. However, for languages structurally and lexically closer to English, such as French and Spanish, finding weaknesses becomes significantly harder, leading to higher costs.}
    \label{fig:cost}
    \vspace{-10pt}
\end{figure*}

\section{Experiment}

\subsection{Experiment Overview}
In this section, we conduct a series of experiments to evaluate the effectiveness of our proposed method as well as to explore the cross-lingual weaknesses of multilingual models. Overall, we mainly aim to address the following questions:

\begin{itemize}[nolistsep, leftmargin=*]
    \item \textbf{RQ1:} How effective and efficient is our method in identifying the cross-lingual weaknesses of multilingual models? (\textbf{\S\ref{section:4.2}})
    \item \textbf{RQ2:} Are the identified weaknesses language-specific?  How can we understand the linguistic connection between cross-lingual weaknesses and the languages involved? (\textbf{\S\ref{section:4.3}})
    \item \textbf{RQ3:} Furthermore, to what extent does language-specific fine-tuning enhance cross-lingual performance, and how is the fine-tuning improvement   associate with the relationships of different languages? (\textbf{\S\ref{Section4.4}})
\end{itemize}

\subsection{Cross-Lingual Weakness Identification}
\label{section:4.2}
To answer RQ1, based on the proposed method, we generated initial bilingual pairs using GPT-4o and employed GPT-4o-mini for perturbation generation. Perturbations were translated using the Google Translate API \citep{googletranslateapi}. We then employed \( W \) cost-effective multilingual models in $\mathcal{M}$ for LLM-based simulation to generate a set of candidates over 6,000 question pairs spanning 16 languages. These pairs are then used to evaluate the performance of 10 different models. Detailed experimental settings and parameter configurations are provided in \autoref{appendix:experiment_details}.

\begin{figure}[t]
    \centering
    \includegraphics[width=\linewidth]{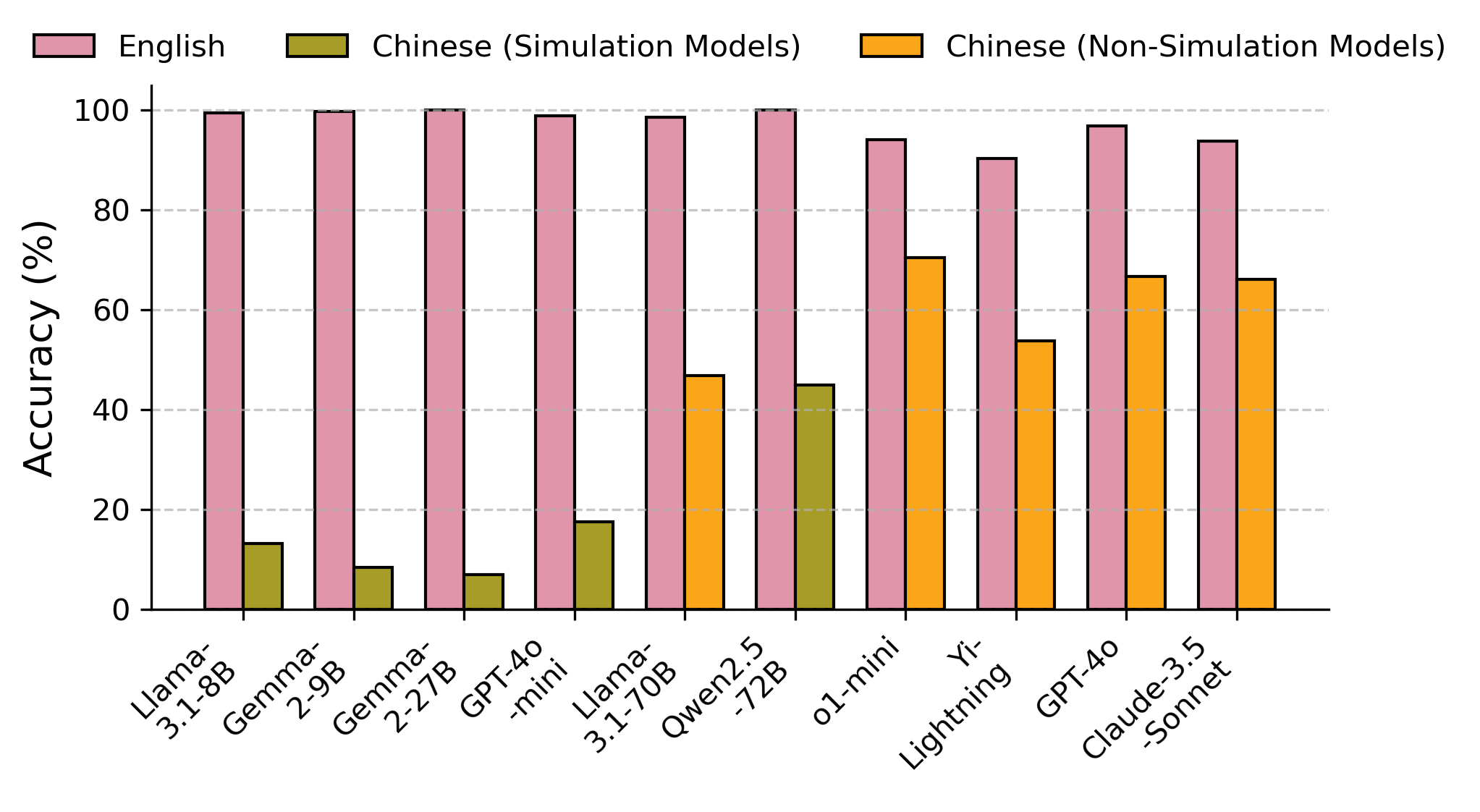} \vspace{-0.2in}
    \caption{Performance of LLMs on our generated English-Chinese pairs. Even smaller models like Gemma-2-9B and Llama-3.1-8B achieve perfect accuracy in English, while more than half of the models score below 50\% in Chinese. Despite their strong multilingual capabilities, GPT-4o and Claude-3.5-sonnet still exhibit over a 30\% accuracy drop compared to English.}
    \label{fig:Chinese_results}
    \vspace{-1em}
\end{figure}

\textbf{Our method effectively identifies cross-lingual weaknesses even in state-of-the-art models.} Taking Chinese as an example, we evaluated all models on our generated English-Chinese pairs and found that their accuracy dropped by nearly 60\% on average when switching from English to Chinese, as shown in \autoref{fig:Chinese_results}. Notably, even the smallest models achieved perfect accuracy on English tasks (i.e., they have mastered the most knowledge of answering the questions), whereas the most advanced model, GPT-4o, still exhibited a substantial accuracy drop of nearly 30\% in Chinese. Similar performance gaps were observed across other languages, as presented in Appendix~\ref{appendix:experiment_procedures}.

\begin{figure*}[t]
    \centering
    \includegraphics[width=\linewidth]{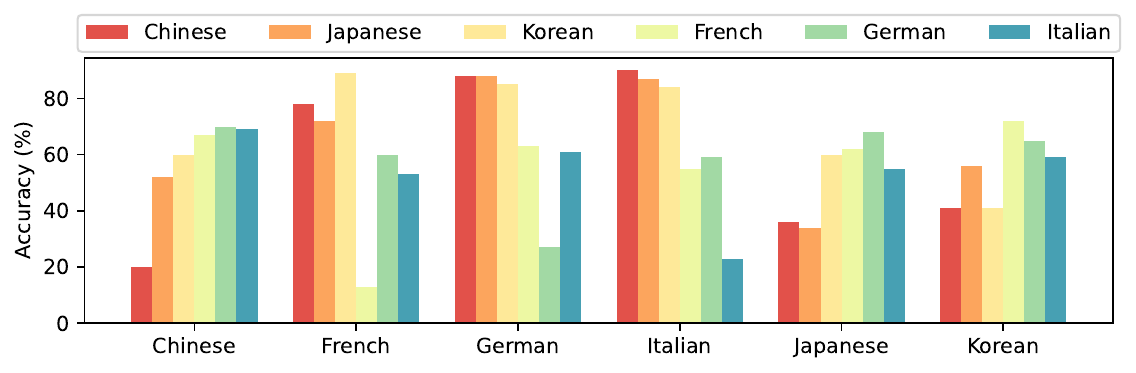}\vspace{-1em}
    \caption{Accuracy of GPT-4o-mini on expanded bilingual pairs (Asian and European language families). The red bar represents accuracy for pairs expanded from Chinese seed pairs, while other colors show results for pairs expanded from other seed language pairs within these families.}
    \label{fig:4o-mini}
    \vspace{-1em}
\end{figure*}

As shown in \autoref{fig:models_languages}, the accuracy drops across 16 languages highlight the cross-lingual performance gaps. Even Claude-3.5-sonnet experienced over 20\% accuracy loss in most languages. This starkly demonstrates the effectiveness of our method in identifying cross-lingual weaknesses in even state-of-the-art multilingual models.

Moreover, from \autoref{fig:Chinese_results}, we can observe that the models used for simulation typically exhibit greater accuracy degradation. By varying the models in $\mathcal{M}$ for LLM-based simulation, we can discover specific cross-lingual weaknesses in any given LLMs. To investigate this, we replaced Gemma-2-27B and Qwen2.5-72B with GPT-4o in our simulation framework. A comparison between \autoref{fig:Chinese_results} and \autoref{fig:4o} reveals that: Qwen2.5-72B and Gemma-2-27B show minor accuracy improvements after being removed from the simulation models, GPT-4o—despite being a top-tier multilingual model—suffers a sharp 58\% accuracy drop.

\textbf{Our method enables the cost-effective identification of cross-lingual weaknesses.} We evaluated the cost of generating bilingual pairs and analyzed the conversion rate for each language—i.e., the proportion of bilingual pairs successfully generated from an original English question—as illustrated in \autoref{fig:cost}. For most languages, the average cost of identifying a question that exposes a model's cross-lingual weaknesses is as low as \$0.05.

Interestingly, for most languages, the cost of generating pairs is significantly lower, compared to the specific languages like French, Spanish, Italian, and German. This discrepancy can be explained by the greater linguistic similarities of these languages to English, particularly in terms of script, vocabulary, and grammatical structures \citep{schepens2012distributions,gnanadesikan2017towards}. For languages that are structurally closer to English, models tend to perform at levels more comparable to their English proficiency \citep{conneau2019unsupervised,pires2019multilingual}, which makes it more challenging to uncover their cross-lingual weaknesses. A more detailed analysis of how linguistic relationships affect cross-lingual model performance is provided in \autoref{section:4.3}.

\textbf{Our search framework significantly outperforms baseline approaches.} We compared our beam search method with two baseline approaches: No Perturbation (NP) and Direct Perturbation (DP). In NP, we directly translate the English questions to target languages without any perturbation, while in DP, we apply perturbations through prompts following the template in \autoref{appendix:prompt_template} directly, without search. Using models in $\mathcal{M}$ for simulation, we identify questions where models perform well in English but fail in target languages. As shown in \autoref{tab:conversion_rates}, our framework consistently achieves substantially higher conversion rates across all evaluated languages compared to both baselines.

\begin{table}[t]
    \centering
    \small
    \caption{Comparison of conversion rates across different languages. \textbf{NP} (No Perturbation) refers to direct translation without perturbations, while \textbf{DP} (Direct Perturbation) applies perturbations without search.} \vspace{-1em}
    \setlength{\tabcolsep}{12pt}
    \label{tab:conversion_rates}
    \begin{tabular}{lccc}
    \toprule
    \textbf{Language} & \textbf{NP} & \textbf{DP} & \textbf{Ours} \\
    \midrule
    Chinese & 0.000 & 0.036 & 0.431 \\
    Japanese & 0.000 & 0.071 & 0.594 \\
    French & 0.000 & 0.018 & 0.132 \\
    German & 0.000 & 0.027 & 0.323 \\
    \bottomrule
    \end{tabular}
    \vspace{-1em}
\end{table}

\subsection{Linguistic Factors in Cross-lingual Weaknesses}
\label{section:4.3}
To answer RQ2, we first sampled 100 seed bilingual pairs (i.e., English-target language pairs) for each of 16 languages from those generated in \autoref{section:4.2}. For each sampled pair, the target-language portion was translated into the other 15 languages, resulting in a total of 25,600 expanded bilingual pairs. These expanded pairs were then evaluated across six different models, with detailed experimental settings outlined in \autoref{appendix:experiment_details}.

\textbf{The identified cross-lingual weaknesses are not restricted to specific languages and depend on the linguistic relationships.} The evaluation results of GPT-4o-mini on expanded pairs from the Asian language family (Chinese, Japanese, and Korean) and the European language family (French, German, and Spanish) are presented in \autoref{fig:4o-mini}. As observed, the model exhibits a consistent decline in accuracy across these pairs.

A clear pattern emerges when analyzing the expanded pairs. Within the Asian language family, weakness pairs expanded from Chinese, Japanese, or Korean into other Asian languages exhibit substantial and relatively consistent accuracy declines. In contrast, when these Asian seed pairs are expanded into European languages, the accuracy drops are considerably smaller and more variable. A similar trend is observed within the European language family: pairs expanded from French, German, or Spanish into other European languages experience significant and consistent accuracy declines, whereas expansion into Asian languages results in smaller and more varied reductions in accuracy. We hypothesize that these patterns are driven by underlying linguistic relationships. Specifically, the Asian language family exhibits shared cross-linguistic challenges, while the European family follows similar patterns. Consequently, expanded pairs from Chinese seed pairs tend to maintain more weakness in Japanese and Korean, whereas those from French seed pairs lead to increased weakness in German and Italian.


\begin{figure}[t]
    \centering
    \includegraphics[width=\linewidth]{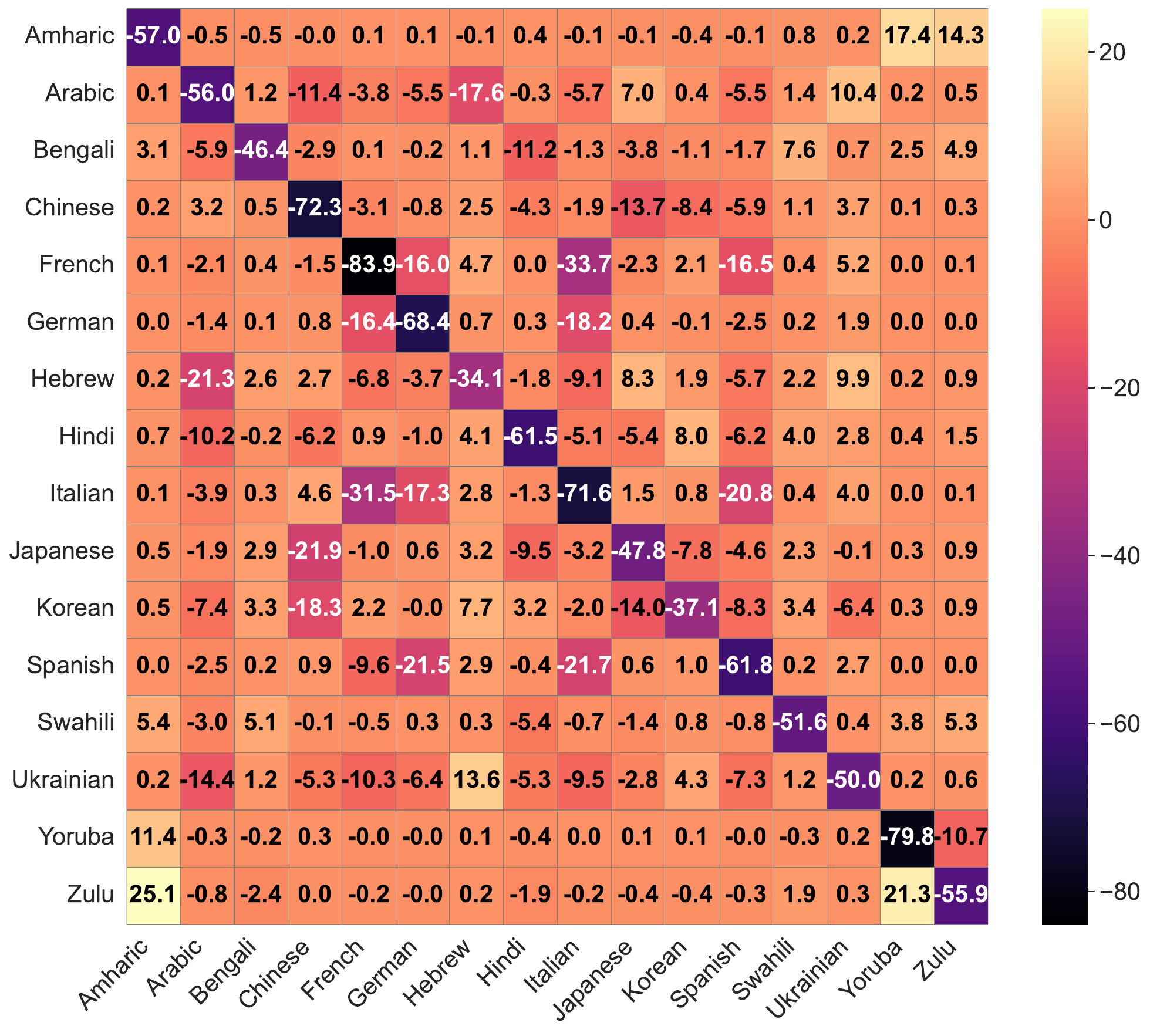}
    \caption{Visualization of RAS \(D_{x, y}\) across 16 languages, highlighting linguistic and cultural proximities. The vertical axis denotes the source language and the horizontal axis denotes the target language. Darker shades of a block indicate stronger retention of shared cross-lingual weaknesses when translating from language \( y \) to language \( x \), signifying a closer linguistic relationship between the two languages.  }
    \label{fig:heatmap}
    \vspace{-1em}
\end{figure}

\textbf{Languages with stronger linguistic affinity tend to exhibit cross-lingual weaknesses in common.} We define the \textbf{Relative Affinity Score (RAS)} \(D_{x, y}\), which measures the linguistic relationships between language \(x\) and language \(y\). The score is computed as:
\[
D_{x, y} = \left( \frac{A_{x, y} - \overline{A}_x}{\overline{A}_x} \right) \cdot \exp\left( c \cdot \left| \overline{A}_y - \overline{A}_x \right| \right)
\]
Here, \(D_{x, y}\) quantifies the linguistic proximity between language \(x\) and \(y\), with lower values indicating a stronger affinity. The term \(A_{x, y}\) represents the model’s accuracy on language \(x\) when using expanded pairs originating from seed language \(y\). The average accuracy on language \(x\) across all seed languages for expanded pairs is denoted by \(\overline{A}_x\). The factor \(\exp\left( c \cdot \left| \overline{A}_y - \overline{A}_x \right| \right)\) scales the score based on the accuracy difference between languages \(x\) and \(y\), where constant \(c\), a negative value, controls the inverse sensitivity of this adjustment.

\begin{figure}[t]
    \centering
    \includegraphics[width=0.9\linewidth]{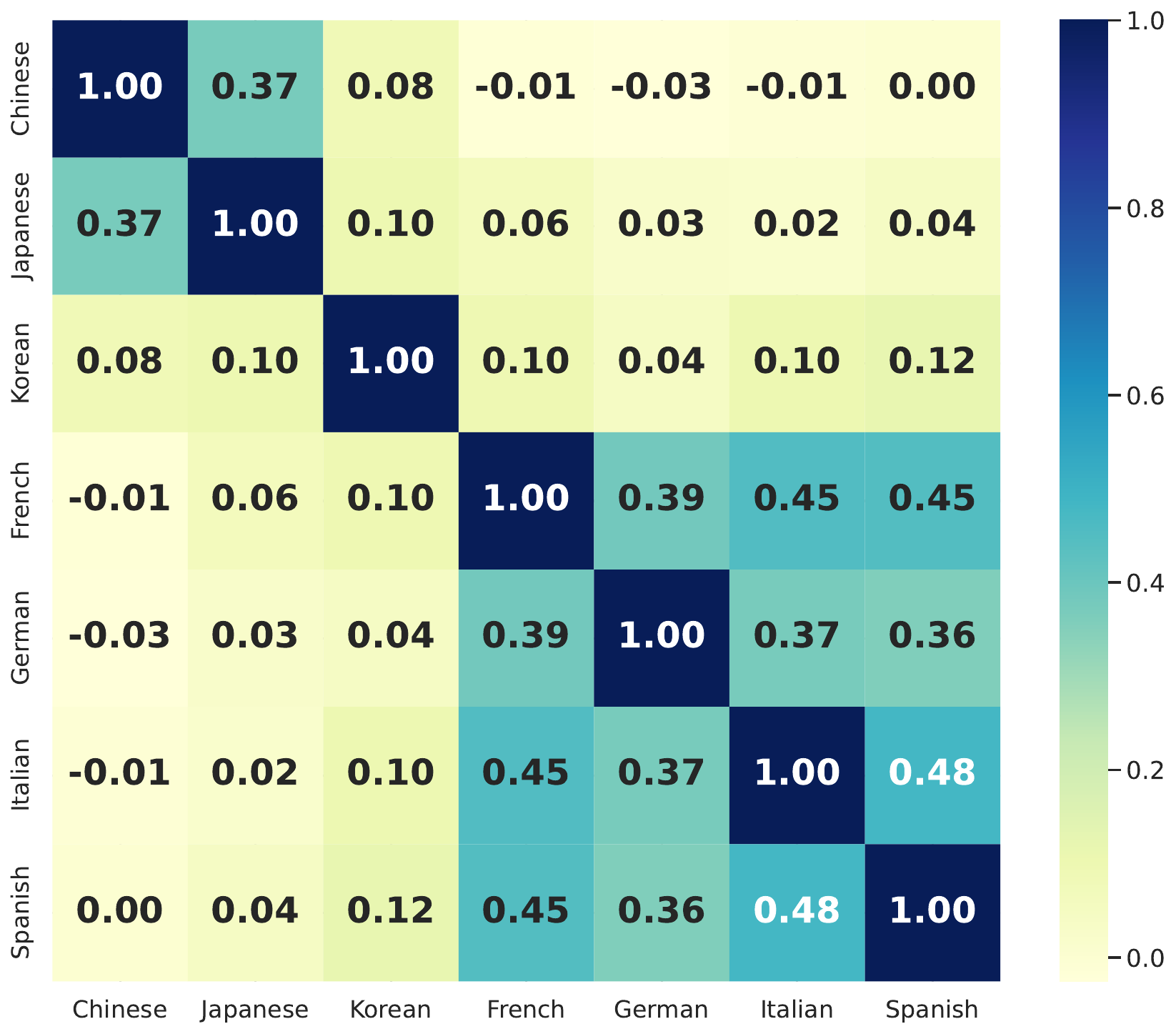}
    \caption{Heatmap of the pairwise cosine distances between the normalized embeddings generated by Llama-3.1-8B for seven English–target language question pairs.}
    \label{fig:embedding_heatmap}
    \vspace{-1em}
\end{figure}

As shown in \autoref{fig:heatmap}, it reveals a clear pattern: lower RAS values \(D_{x,y}\) are predominantly observed for language pairs (\(x, y\)) with linguistic and cultural proximities. This observation strongly supports our hypothesis that languages with closer linguistic ties tend to share cross-lingual weaknesses.

\begin{table*}[t]
  \centering
  \small
  \renewcommand{\arraystretch}{1}
  \setlength{\tabcolsep}{7pt}
  \caption{Performance comparison of Phi-3.5-Mini and Llama-3.1-8B after SFT and DPO on French and Chinese datasets. The table shows evaluation results on various evaluation languages (EL), with the Asian language group highlighted in blue.  Performance differences (Diff.) are shown compared to the original model (Orig.).  "S Enh." represents the model enhanced by SFT, and "D Enh." represents the model enhanced by simulated DPO. Due to space limitations, performance of Gemma-2-9B and Qwen2.5-7B are presented in \autoref{tab:ft_full_dpo_supp}.}
  \label{tab:ft_one_dpo}
  \begin{tabular}{lcccccccccc}
    \toprule
    & & & \multicolumn{4}{c}{\textbf{French Fine-Tuning}} & \multicolumn{4}{c}{\textbf{Chinese Fine-Tuning}} \\
    \cmidrule(lr){4-7} \cmidrule(lr){8-11}
    \multirow{-2}{*}{\textbf{Model}} & \multirow{-2}{*}{\textbf{EL}} & \multirow{-2}{*}{\textbf{Orig.}} & \textbf{S Enh.} & \textbf{D Enh.} & \textbf{S Diff.} & \textbf{D Diff.} & \textbf{S Enh.} & \textbf{D Enh.} & \textbf{S Diff.} & \textbf{D Diff.} \\
    \midrule
    \multirow[c]{6}{*}{\textbf{Phi-3.5-Mini}} & French   & 0.196     & --    & --    & -- & -- & 0.397 & 0.425 &   0.202  & 0.229 \\
     & German   & 0.208          & 0.466 &  0.488 & 0.258 & 0.258 & 0.441 & 0.465&  0.233  & 0.257 \\
     & Spanish  & 0.248         & 0.467 & 0.495 & 0.219 & 0.247 & 0.521 & 0.550 & 0.273 & 0.302  \\
     & \cellcolor{lightblue}Chinese & \cellcolor{lightblue}0.199 & \cellcolor{lightblue}0.345 & \cellcolor{lightblue}0.330 & \cellcolor{lightblue}0.146 & \cellcolor{lightblue}0.131  & \cellcolor{lightblue}--    & \cellcolor{lightblue}--    & \cellcolor{lightblue}--   & \cellcolor{lightblue}--  \\
     & \cellcolor{lightblue}Japanese& \cellcolor{lightblue}0.127 & \cellcolor{lightblue}0.306 & \cellcolor{lightblue}0.325 & \cellcolor{lightblue}0.178 & \cellcolor{lightblue}0.198 & \cellcolor{lightblue}0.478 & \cellcolor{lightblue}0.510 &  \cellcolor{lightblue}0.350 &  \cellcolor{lightblue}0.383  \\
     & \cellcolor{lightblue}Korean  & \cellcolor{lightblue}0.086   & \cellcolor{lightblue}0.246 &  \cellcolor{lightblue}0.265 & \cellcolor{lightblue}0.160 & \cellcolor{lightblue}0.179 & \cellcolor{lightblue}0.458 & \cellcolor{lightblue}0.492 &  \cellcolor{lightblue}0.373  & \cellcolor{lightblue}0.406 \\
    \midrule
    \multirow[c]{6}{*}{\textbf{Llama-3.1-8B}} & French   & --    & --    & --    & --    & --    & 0.506 & 0.518 & 0.343 & 0.355 \\
     & German   & 0.189 & 0.388 & 0.395 & 0.199 & 0.206 & 0.494 & 0.502 & 0.304 & 0.312 \\
     & Spanish  & 0.145 & 0.409 & 0.418 & 0.264 & 0.273 & 0.438 & 0.445 & 0.293 & 0.300 \\
     & \cellcolor{lightblue}Chinese & \cellcolor{lightblue}0.289 & \cellcolor{lightblue}0.453 & \cellcolor{lightblue}0.445 & \cellcolor{lightblue}0.164 & \cellcolor{lightblue}0.156 & \cellcolor{lightblue}--    & \cellcolor{lightblue}--    & \cellcolor{lightblue}--   & \cellcolor{lightblue}--    \\
     & \cellcolor{lightblue}Japanese& \cellcolor{lightblue}0.124 & \cellcolor{lightblue}0.357 & \cellcolor{lightblue}0.368 & \cellcolor{lightblue}0.232 & \cellcolor{lightblue}0.243 & \cellcolor{lightblue}0.561 & \cellcolor{lightblue}0.570 & \cellcolor{lightblue}0.436 & \cellcolor{lightblue}0.445 \\
     & \cellcolor{lightblue}Korean  & \cellcolor{lightblue}0.143 & \cellcolor{lightblue}0.320 & \cellcolor{lightblue}0.328 & \cellcolor{lightblue}0.178 & \cellcolor{lightblue}0.185 & \cellcolor{lightblue}0.504 & \cellcolor{lightblue}0.515 & \cellcolor{lightblue}0.362 & \cellcolor{lightblue}0.373 \\
    \bottomrule
  \end{tabular}\
  \vspace{-1em}
\end{table*}

We further investigated the linguistic basis of these cross-lingual weaknesses by analyzing the embedding space of seed bilingual pairs. Specifically, we embedded cross-lingual weaknesses identified in \autoref{section:4.2} for the Asian and European language families. As shown in  \autoref{fig:embedding_tsne}, visualizing these embeddings via t-SNE revealed a clustering effect: weaknesses from the same family clustered together. This observation was corroborated by the cosine distance matrix, as presented in \autoref{fig:embedding_heatmap}, which showed significantly smaller embedding distances within the Asian and European families compared to distances between families. 

\textbf{Cross-lingual weaknesses correlate with specific subject domains.} To further investigate cross-lingual weaknesses, we categorized the identified bilingual pairs into six subject domains: Science \& Technology, History \& World Events, Society \& Culture, Arts \& Literature, Geography \& Environment, and General Knowledge. The distribution of these weaknesses across categories, broken down by language, is detailed in \autoref{tab:category}. Notably, lower-resource languages such as Amharic, Arabic, and Yoruba exhibited significantly more errors in the Science \& Technology domain compared to most languages. Conversely, higher-resource languages like Chinese, Spanish, and German demonstrated stronger performance in this area. Interestingly, Chinese showed a distinct weakness in Society \& Culture, while Korean displayed comparatively weaker performance in Geography \& Environment.

\subsection{Cross-lingual Fine-tuning}
\label{Section4.4}

To answer RQ3, we designed a fine-tuning experiment to explore whether language-specific fine-tuning preferentially enhances performance on linguistically similar languages. We focused on two language families identified as linguistically proximate in our earlier analysis: the Asian language family (Chinese, Japanese, and Korean) and the European language family (French, German, and Spanish). Using the Chinese and French seed pairs identified in \autoref{section:4.2}, we performed both supervised fine-tuning (SFT) and Direct Preference Optimization (DPO) \citep{rafailov2023direct} on several LLMs: Phi-3.5-Mini, Gemma-2-9B, Llama-3.1-8B, Qwen2.5-7B. Separate fine-tuning runs were conducted using both Chinese and French portion in seed pairs. Subsequently, we evaluated the performance of these fine-tuned models across the other languages in two language families. This experiment aimed to investigate if fine-tuning on a specific language leads to greater performance gains in linguistically related languages.

As shown in \autoref{tab:ft_one_dpo} and \autoref{tab:ft_full_dpo_supp}, the evaluation results reveal a consistent trend: fine-tuning on Chinese significantly improves performance in Japanese and Korean, while its impact on European languages is comparatively smaller. Similarly, fine-tuning on French enhances performance in related European languages like German and Spanish but has a weaker effect on Asian languages. This pattern holds across both SFT and DPO fine-tuning, indicating that linguistic proximity, rather than the fine-tuning method, primarily drives cross-lingual knowledge transfer. These findings suggest that current LLMs inherently capture linguistic relationships, facilitating more effective transfer between closely related languages.

\section{Discussion}
\label{sec:discussion}

Our investigation into cross-lingual weaknesses underscores several critical aspects for both understanding current LLM limitations and paving the way for future improvements.

First, the integrity of our findings hinges on the \textbf{quality of translation} in bilingual question pairs. If semantic equivalence between the English source and target language question is not rigorously maintained, observed performance drops could be mistakenly attributed to the model's cross-lingual deficiencies rather than translation artifacts. To mitigate this, we employed LLMs for both initial translation and semantic verification, a widely adopted practice in multilingual research \citep{lin2024investigating,ye2024justice}. The efficacy of this approach was further corroborated through human evaluation, whose methodology and results are presented in \autoref{appendix:human}. The evaluation confirmed that most generated pairs exhibit high translational fidelity. As multilingual capabilities of LLMs continue to advance, developing more sophisticated and reliable translation and semantic checking components will be instrumental in refining the precision with which cross-lingual weaknesses are identified and analyzed.

Second, to provide a richer, more nuanced understanding beyond aggregate statistics, we have compiled an extensive set of \textbf{case studies}. These qualitative examples, detailed in \autoref{appendix:case}, illustrate the diverse nature of cross-lingual pitfalls encountered by various models across different languages. They showcase specific failure modes, such as misinterpretation of nuanced phrasing, incorrect entity mapping, or breakdowns in reasoning when faced with linguistic structures that differ significantly from English. These case studies offer valuable material for researchers seeking to conduct in-depth analyses of specific cross-lingual phenomena or to understand the particular challenges faced by individual models or language families.

Finally, the identification of these cross-lingual weaknesses is not merely an academic exercise but offers substantial \textbf{potential for enhancing the multilingual capabilities of LLMs}. Our methodology serves as a diagnostic tool, pinpointing specific areas where LLMs falter, thereby guiding targeted interventions. For instance, the weaknesses uncovered can inform more focused \textbf{fine-tuning strategies}, concentrating efforts on language pairs or specific linguistic constructions where models demonstrate pronounced deficiencies, potentially leveraging the subject domain categorizations (as shown in \autoref{tab:category}) to further refine this targeting. Furthermore, the challenging cross-lingual examples generated by our method can be invaluable for \textbf{augmenting pre-training and instruction-tuning datasets} \citep{huang2024datagen}. By enriching training corpora with instances that expose known weaknesses, we can proactively address data imbalances or representational gaps that contribute to these performance discrepancies. Lastly, these targeted examples are well-suited for \textbf{continual learning or adaptive training paradigms}, enabling models to iteratively strengthen their cross-lingual understanding and reasoning in precisely the areas where they have been shown to be vulnerable. In essence, a systematic approach to uncovering weaknesses, such as the one proposed, is a crucial first step towards building more robust multilingual LLMs.

\section{Conclusion}
In this study, we proposed an efficient beam search with LLM-based simulation to identify cross-lingual weaknesses in LLMs, generating a 16-language dataset that exposed performance gaps even in state-of-the-art models. Our findings highlight linguistic relationships as key to shared vulnerabilities and fine-tuning benefits, emphasizing the need to consider linguistic nuances in developing truly multilingual LLMs.

\section*{Limitations}

While our methodology demonstrates effectiveness in identifying cross-lingual weaknesses, several avenues for future refinement exist. First, the current study's scope, while covering a diverse set of languages, is not fully comprehensive. A more complete picture of cross-lingual consistency in LLMs would require extending our analysis to a broader range of languages, particularly those with limited resources or significantly different structural characteristics. Relatedly, although we employ LLM-based semantic checks to ensure the semantic equivalence of our bilingual question pairs, subtle nuances arising from cultural context or idiomatic expressions might still introduce minor biases. Finally, our core approach of iteratively adding perturbations is effective at revealing weaknesses related to complexity. However, this strategy may be less sensitive to identifying those vulnerabilities that manifest in very short, concise prompts. Consequently, investigating complementary techniques specifically designed for such cases would enhance the overall robustness of our framework.

\section*{Ethics Statement}
This research adheres to ethical standards in AI research and development. Our methodology is designed to identify and understand cross-lingual weaknesses in LLMs to improve their multilingual capabilities. We recognize the potential for bias within LLMs, particularly across different languages and cultural contexts. Our language selection was carefully considered to ensure diversity, encompassing both high-resource and lower-resource languages. All generated content and model outputs were scrutinized for potential biases. No personally identifiable information was collected or used. This work is intended to promote inclusivity and fairness in the development of multilingual LLMs. The findings are shared with the research community to foster further investigation and the mitigation of cross-lingual weaknesses in LLMs.

\bibliography{custom}

\clearpage
\appendix

\section{Related Work}

\subsection{LLM Evaluation}

Significant efforts have been devoted to evaluating the capabilities of LLMs across a wide range of domains. These evaluations include traditional NLP tasks such as sentiment analysis \citep{zhang2023sentiment,wan2025cognitive} and translation \citep{yao2023benchmarking,zhang2023prompting}, as well as mathematical reasoning \citep{hendrycks2021measuring,liu2024mathbench}, scientific and domain-specific question answering \citep{xu2025gta,luo2023nqe,zhou2023path,chen2025unveiling,song2025injecting}, and coding skills \citep{chen2021evaluating,jain2024livecodebench}. Evaluations have also extended into specialized domains such as chemistry \citep{chen2025unveiling}, medicine \citep{xie2025medtrinity25mlargescalemultimodaldataset}, and geolocation reasoning \citep{song2025geolocation}. In the area of cybersecurity, efforts have been made to assess LLMs’ ability to detect software vulnerabilities \citep{liu2024vuldetectbenchevaluatingdeepcapability}. Beyond task performance, growing attention has been paid to trustworthiness \citep{sun2024trustllm,chujie2024honestllm}, including robustness to spurious correlations \citep{liu2025adversarialcooperativerationalizationrisk}, resilience to textual perturbations \citep{wang2025wordformmattersllms}, and defense against jailbreak attacks \citep{gao2024shapingsafetyboundariesunderstanding}. Comprehensive benchmarks and investigations have been proposed to systematically assess these aspects \citep{huang2025trustworthiness,wang2025trusteval}. General-purpose benchmarks like MMLU \citep{hendryckstest2021,hendrycks2021ethics} continue to serve as a foundation for evaluating broad LLM capabilities.

In this study, we select a subset of English questions from five widely used question-answering datasets: CommonsenseQA\citep{naveed2023comprehensive}, ARC\citep{clark2018think}, MMLU\citep{hendryckstest2021,hendrycks2021ethics}, SciQ\citep{welbl2017crowdsourcing}, and TruthfulQA\citep{lin2021truthfulqa}. These datasets evaluate models on common sense reasoning, mathematical problem-solving, scientific knowledge, and various other skills. We use these questions as the foundation for generating our own dataset.

\subsection{Cross-lingual Capablity of LLMs.}

The cross-lingual capabilities of LLMs have become a central focus in NLP research. Multitask finetuning (MTF) has proven effective for enhancing cross-lingual generalization, as shown by \citet{muennighoff2022crosslingual}, where finetuning multilingual models like BLOOM and mT5 on English tasks enabled zero-shot task transfer to other languages. Beyond MTF, cross-lingual prompting techniques such as chain-of-thought (CoT) prompting \citep{qin2023cross} improve reasoning accuracy by aligning representations and employing task-specific solvers. Other approaches, including cross-lingual knowledge editing \citep{wang2023cross}, entity-based data augmentation \citep{yamada2024leia} and cross-lingual knowledge aggregator\citep{huang20241+}, have been proposed to enhance adaptation and infuse models with cross-lingual knowledge.

Evaluation has also gained attention, with frameworks like the Cross Lingual Auto Evaluation (CIA) Suite \citep{doddapaneni2024cross} addressing challenges in assessing multilingual model outputs. However, many MTF studies remain English-centric \citep{muennighoff2022crosslingual}, and prompting techniques \citep{qin2023cross} may struggle with diverse linguistic structures. While methods like adapter merging \citep{zhao2024adamergex} and continual pre-training \citep{fujii2024continual} aim to enhance language transfer, systematic investigation into multilingual LLM weaknesses across diverse languages remains limited. Additionally, while studies probe cross-lingual alignment during pre-training \citep{wang2024probing,liu2023alignbench} and its importance \citep{hammerl2024understanding}, a quantifiable measure of linguistic relationships affecting cross-lingual transfer is absent. 

Our work builds on these foundations by systematically identifying and analyzing cross-lingual weaknesses in LLMs across 16 diverse languages. By introducing a novel metric to quantify linguistic relationships based on observed performance, we offer deeper insights into how linguistic relation impacts model behavior.

\definecolor{headerbg}{RGB}{255, 183, 77}  
\definecolor{rowgray}{RGB}{255, 236, 209}  
\definecolor{rowblue}{RGB}{255, 247, 230}  
\definecolor{darkgreen}{RGB}{0, 150, 0}  
\begin{table*}[htbp]
\centering
\small
\renewcommand{\arraystretch}{1}
\setlength{\tabcolsep}{2.4mm}
\caption{Models used in our experiments along with their versions, organizations, licenses, and purposes. \textit{Eval}: Model used for evaluation; \textit{FT}: Model used for fine-tuning.}
\label{tab:all_models}
\rowcolors{2}{rowblue}{rowgray}  
\begin{tabular}{lccccc}
    \toprule[1.5pt]
    \rowcolor{headerbg}
    \textcolor{white}{\textbf{Model}} & 
    \textcolor{white}{\textbf{Version}} & 
    \textcolor{white}{\textbf{Organization}} & 
    \textcolor{white}{\textbf{License}} & 
    \textcolor{white}{\textbf{Eval}} & 
    \textcolor{white}{\textbf{FT}} \\
    \midrule[0.8pt]
    GPT-4o-mini       & gpt-4o-mini-2024-07-18       & OpenAI      & Proprietary            & \textcolor{darkgreen}{\checkmark} & \\
    GPT-4o            & gpt-4o-2024-08-06            & OpenAI      & Proprietary            & \textcolor{darkgreen}{\checkmark} & \\
    Gemma-2-9B        & Gemma-2-9B-it                & Google      & Gemma License          & \textcolor{darkgreen}{\checkmark} & \textcolor{darkgreen}{\checkmark} \\
    Gemma-2-27B       & Gemma-2-27B-it               & Google      & Gemma License          & \textcolor{darkgreen}{\checkmark} & \\
    Llama-3.1-8B      & Meta-Llama-3.1-8B-Instruct   & Meta        & Llama 3.1 Community    & \textcolor{darkgreen}{\checkmark} & \textcolor{darkgreen}{\checkmark} \\
    Llama-3.1-70B     & Meta-Llama-3.1-70B-Instruct  & Meta        & Llama 3.1 Community    & \textcolor{darkgreen}{\checkmark} & \\
    Yi-Lightning      & Yi-Lightning                 & 01 AI       & Proprietary            & \textcolor{darkgreen}{\checkmark} & \\
    Qwen2.5-7B        & Qwen2.5-7B-Instruct          & Alibaba     & Qwen License           & \textcolor{darkgreen}{\checkmark} & \textcolor{darkgreen}{\checkmark} \\
    Qwen2.5-72B       & Qwen2.5-72B-Instruct         & Alibaba     & Qwen License           & \textcolor{darkgreen}{\checkmark} & \\
    o1-mini           & o1-mini-2024-09-12           & OpenAI      & Proprietary            & \textcolor{darkgreen}{\checkmark} & \\
    Phi-3.5-mini      & Phi-3.5-mini-instruct        & Microsoft   & MIT                    &  & \textcolor{darkgreen}{\checkmark} \\
    Claude-3.5-Sonnet & claude-3-5-sonnet-20241022   & Anthropic   & Proprietary            & \textcolor{darkgreen}{\checkmark} & \\
    \bottomrule[1.5pt]
\end{tabular}
\end{table*}

\section{Experiment Details}
\label{appendix:experiment_details}

\subsection{Experiment Settings}
\label{appendix:experiment_settings}

\textbf{Source dataset.} To create bilingual pairs, we randomly sampled English questions from five commonly used datasets that cover a wide range of model capabilities: ARC, MMLU, CommonsenseQA, TruthfulQA, and SciQ.  The sampling was performed equally across all five datasets.

\textbf{Models.} As detailed in \autoref{tab:all_models}, we utilized five proprietary models: GPT-4o \citep{hurst2024gpt}, GPT-4o-mini \citep{openai2024gpt4omini}, Yi-Lightning \citep{wake2024yi}, Claude-3.5-Sonnet \citep{anthropic2024claude35}, and o1-mini \citep{jaech2024openai}. In addition, we included seven open-weight models: Gemma-2-9B, Gemma-2-27B \citep{gemma_2024}, Qwen2.5-7B, Qwen2.5-72B \citep{qwen2,qwen2.5}, Llama-3.1-8B \citep{meta2024llama31_8b}, Llama-3.1-70B \citep{meta2024llama31_70b}, and Phi-3.5-mini \citep{abdin2024phi}.

\textbf{Hyperparameter settings.} For perturbation generation, we used a temperature of 0.7 to encourage more diverse and creative responses. In the translation, semantic checking, and simulation tasks, the temperature was reduced to 0.001 to ensure stability in the responses. The maximum output length for these tasks was capped at 1,024 tokens. During beam search, we initialized the process with \(W = 4\) bilingual pairs, and the search width was set to \(w = 12\). The search depths were configured to \(d_1 = 4\) and \(d_2 = 6\), respectively. To promote diversity in the generated questions, we set \(r = 3\). The simulation score parameter, \(\gamma\), was set to 2. For the Early Stopping Mechanism, \(\theta_{pot}\) was set to 0.6, and for determining inclusion in the candidate list, \(\theta_{inc}\) was set to 0.8.  We employed \(K = 5\) LLMs for LLM-based simulation. The constant \(C\) used to calculate the Relative Affinity Score was set to -1.

For the fine-tuning experiments, we trained for 4 epochs with a learning rate of 2.0e-4, employing a cosine learning rate scheduler and a warmup ratio of 0.1. The per-device training batch size was 1, with a gradient accumulation of 8 steps. For evaluation, we used 10\% of the training data as a validation set, evaluated every 200 steps, and set the per-device evaluation batch size to 1.

\subsection{Experiment Procedures}
\label{appendix:experiment_procedures}
\begin{figure}[htbp]
    \centering
    \includegraphics[width=\linewidth]{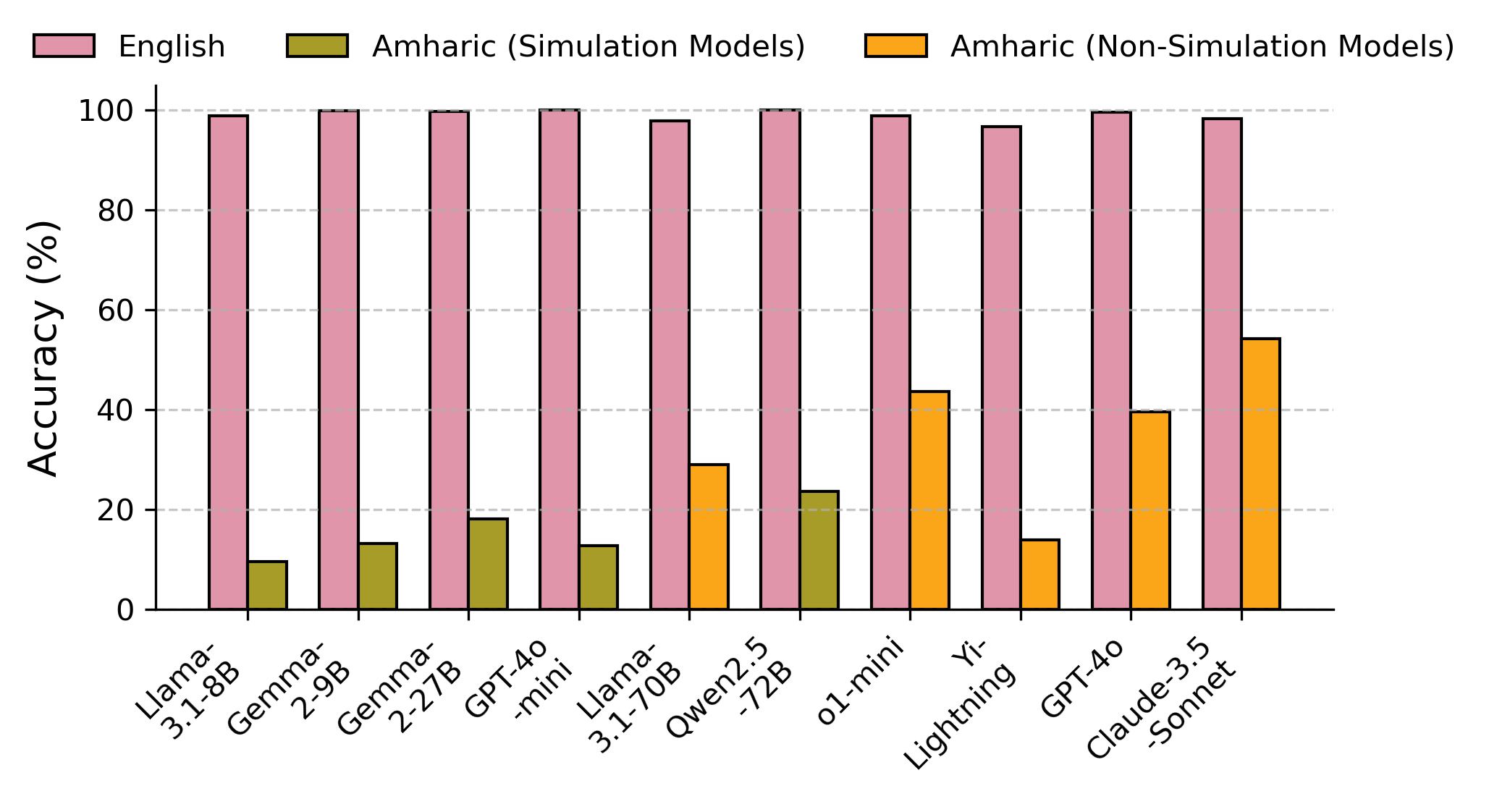}
    \caption{Performance of LLMs on English-Amharic pairs in our candidate list.}
    \label{fig:Amharic_results}
\end{figure}

\begin{figure}[htbp]
    \centering
    \includegraphics[width=\linewidth]{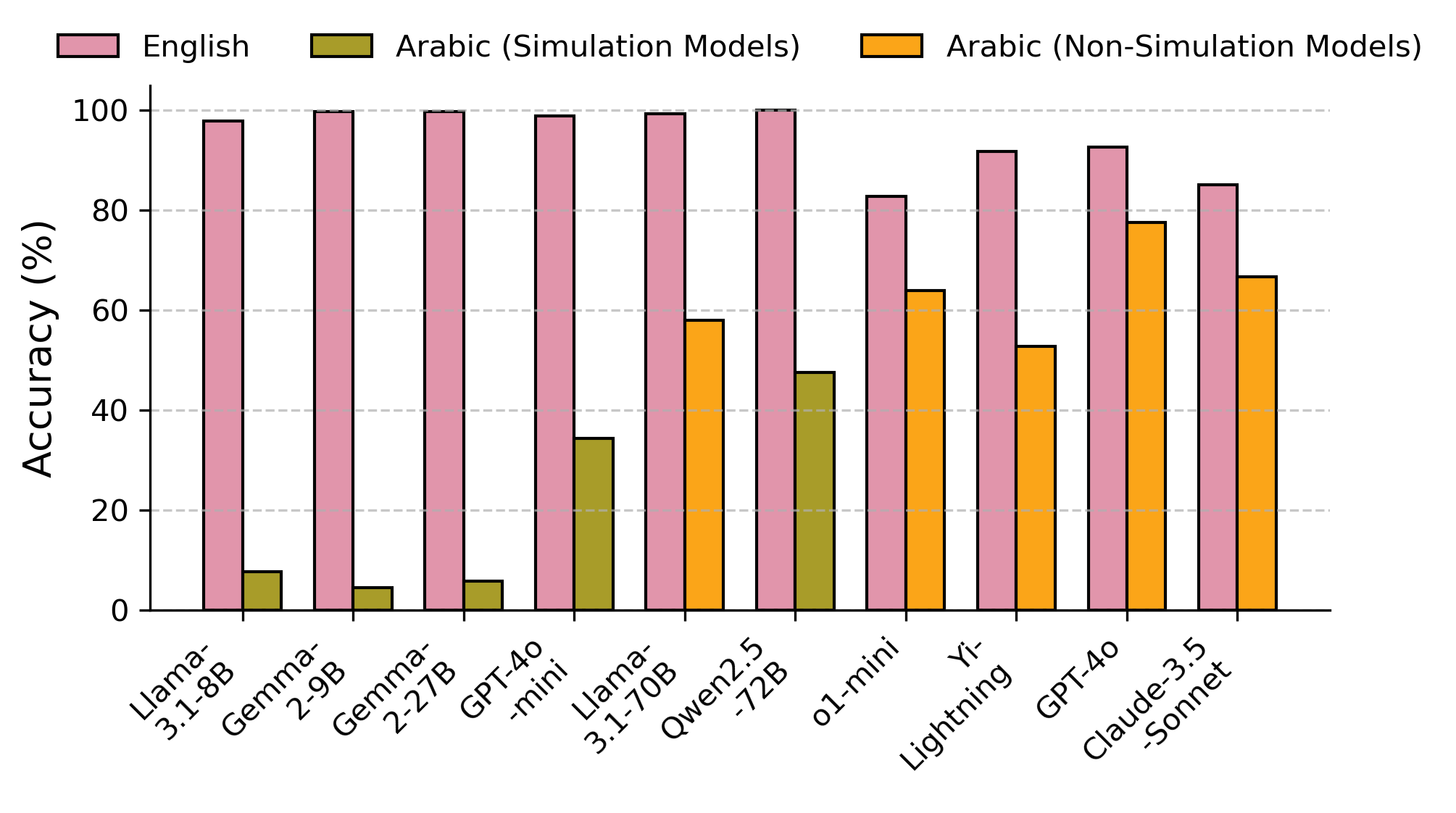}
    \caption{Performance of LLMs on English-Arabic pairs in our candidate list.}
    \label{fig:Arabic_results}
\end{figure}

\begin{figure}[htbp]
    \centering
    \includegraphics[width=\linewidth]{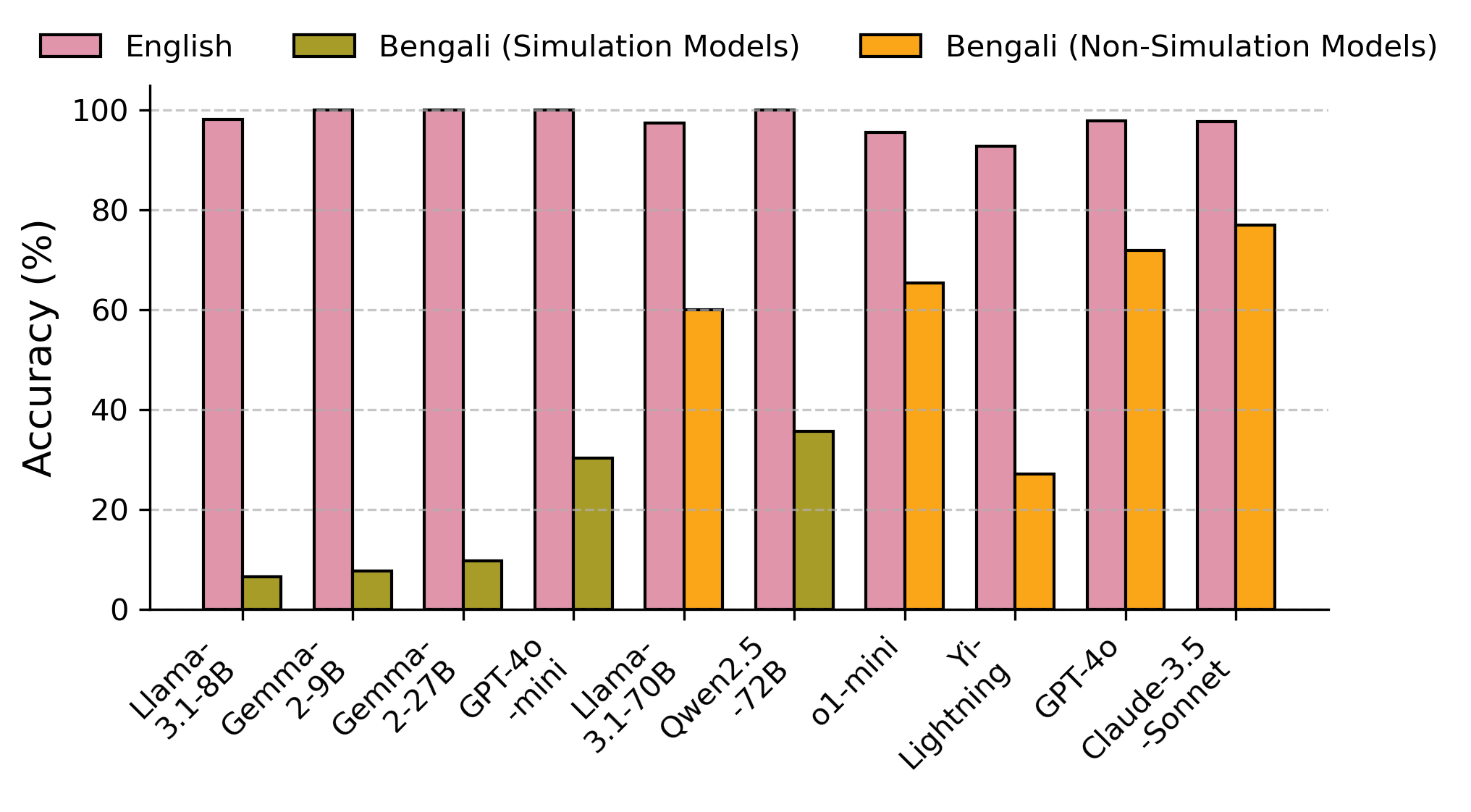}
    \caption{Performance of LLMs on English-Bengali pairs in our candidate list.}
    \label{fig:Bengali_results}
\end{figure}

\begin{figure}[htbp]
    \centering
    \includegraphics[width=\linewidth]{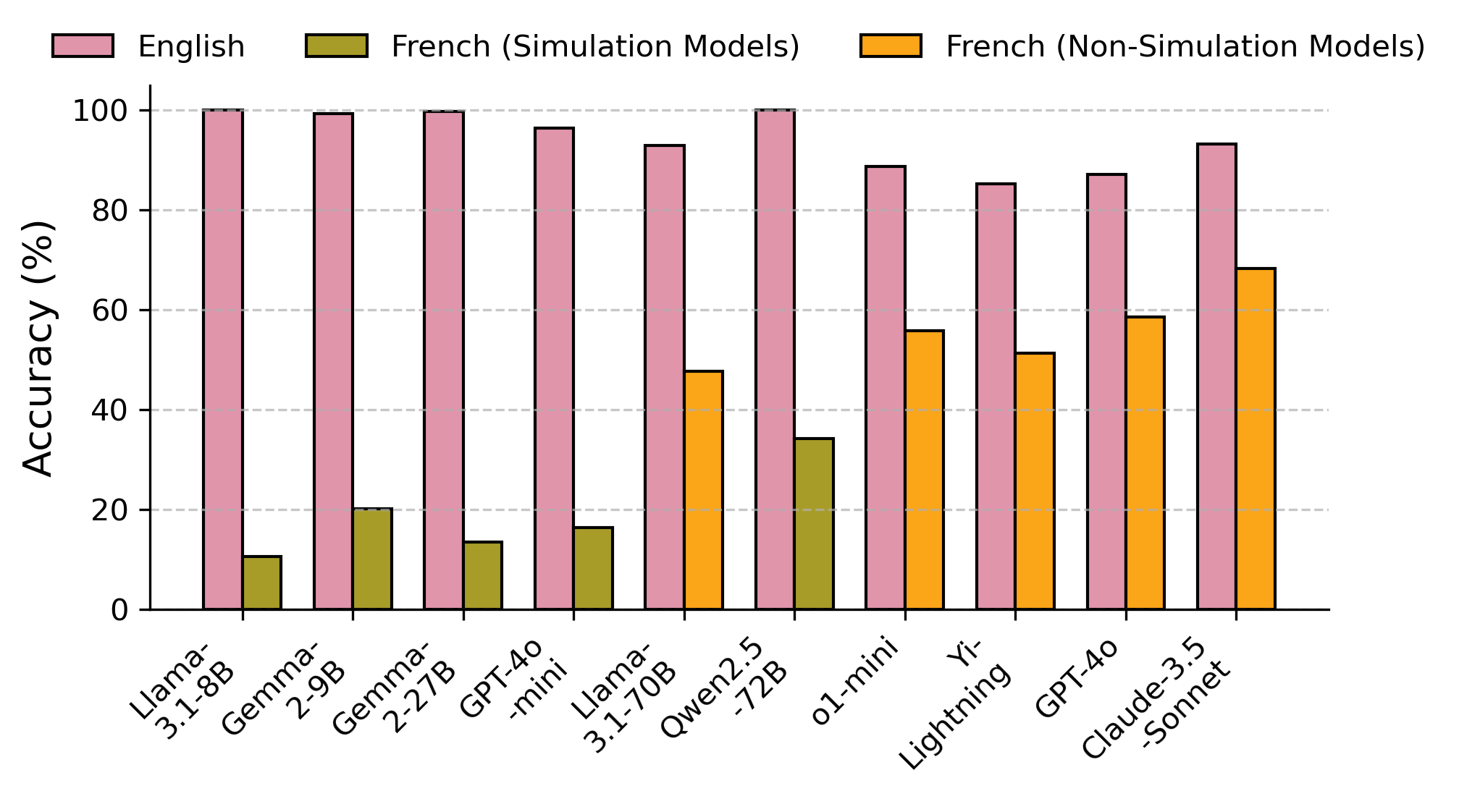}
    \caption{Performance of LLMs on English-French pairs in our candidate list.}
    \label{fig:French_results}
\end{figure}

\begin{figure}[htbp]
    \centering
    \includegraphics[width=\linewidth]{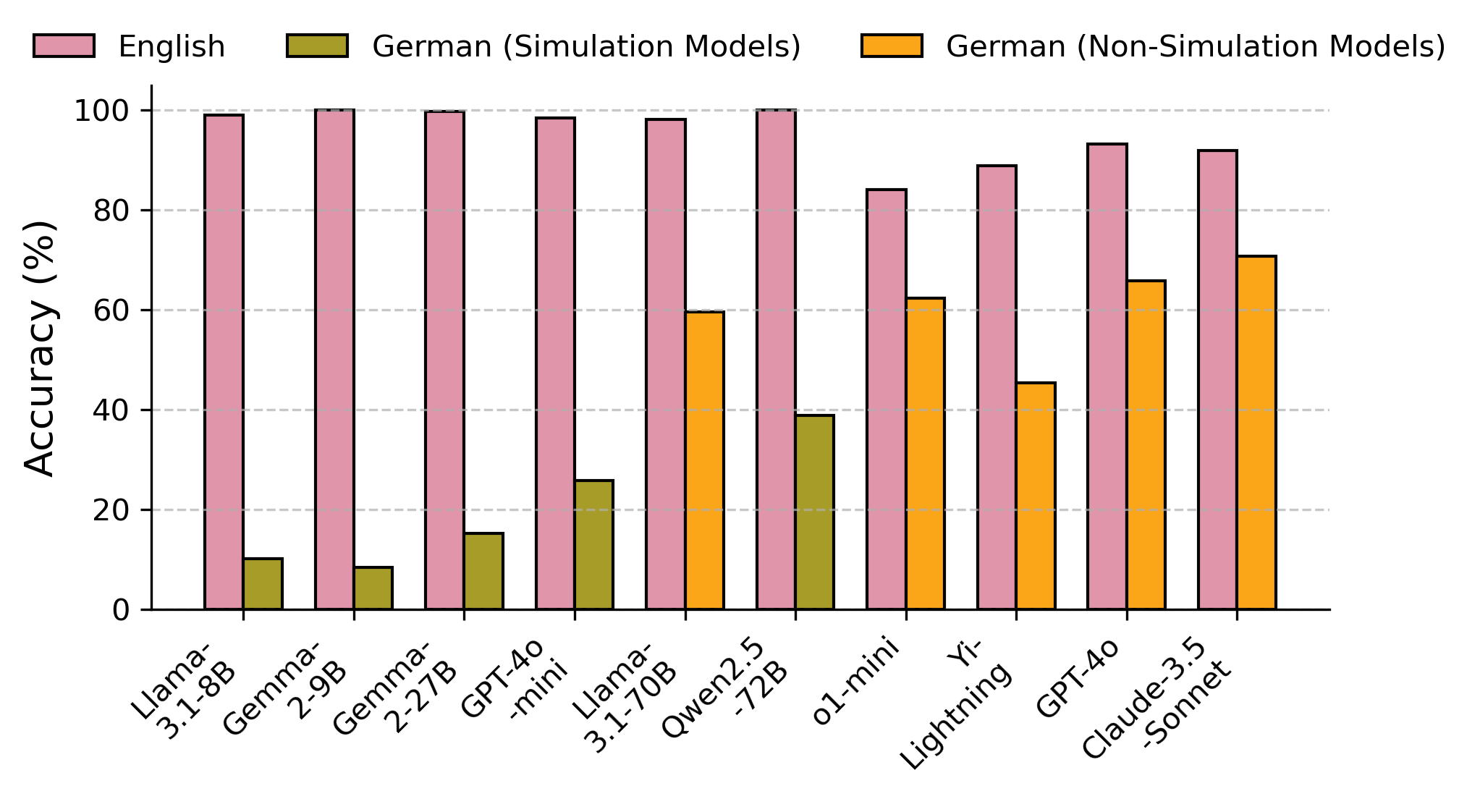}
    \caption{Performance of LLMs on English-German pairs in our candidate list.}
    \label{fig:German_results}
\end{figure}

\begin{figure}[htbp]
    \centering
    \includegraphics[width=\linewidth]{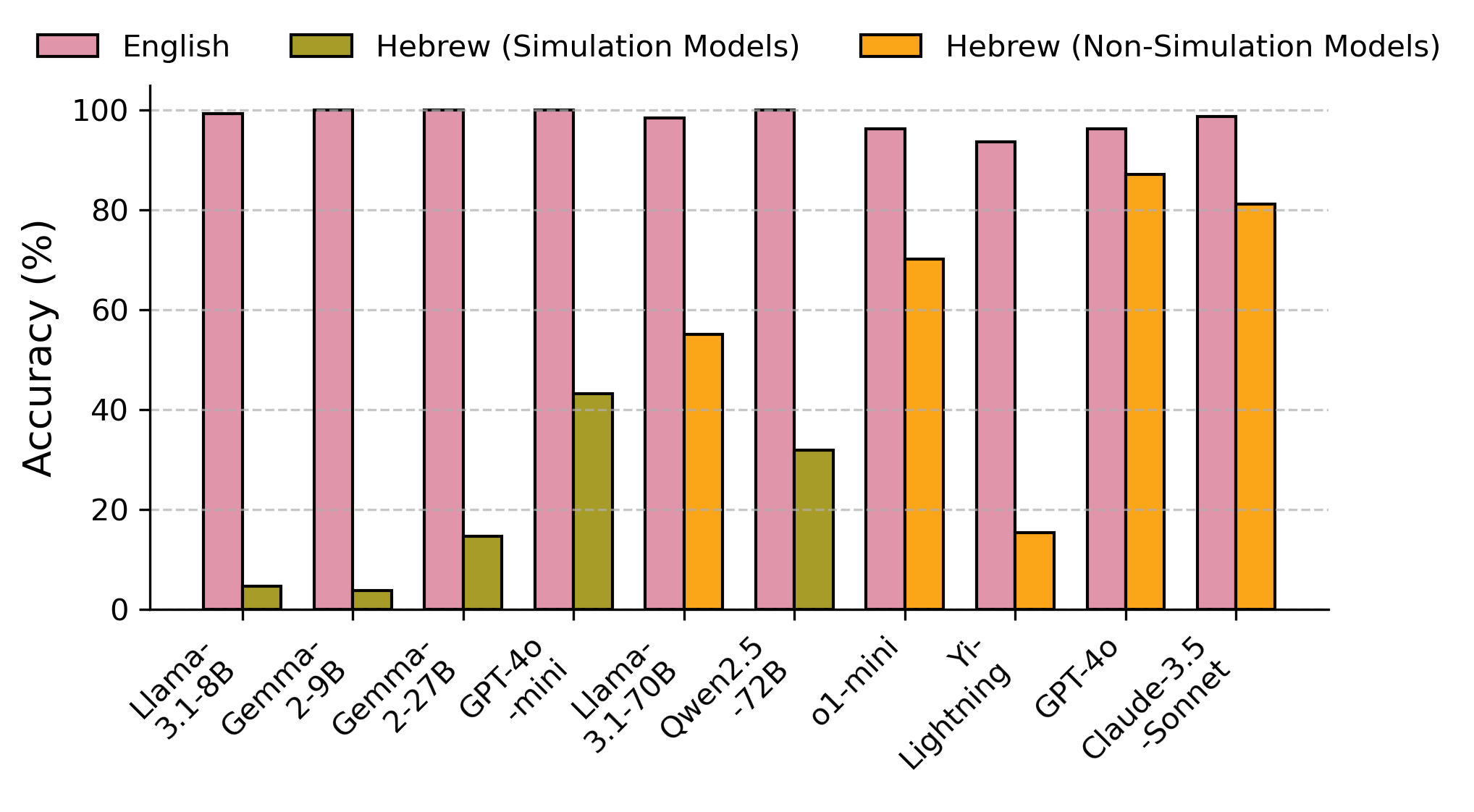}
    \caption{Performance of LLMs on English-Hebrew pairs in our candidate list.}
    \label{fig:Hebrew_results}
\end{figure}

\begin{figure}[htbp]
    \centering
    \includegraphics[width=\linewidth]{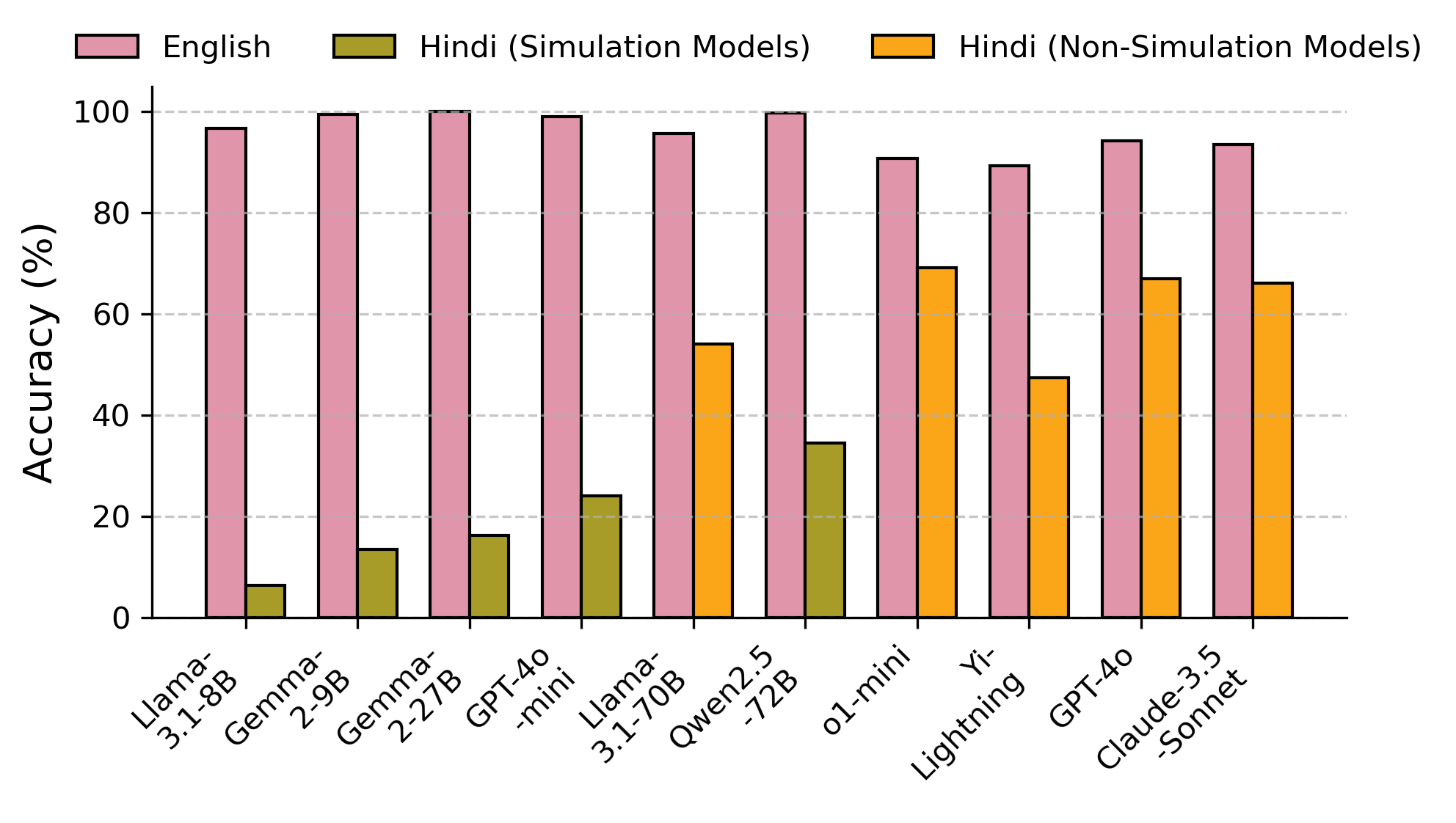}
    \caption{Performance of LLMs on English-Hindi pairs in our candidate list.}
    \label{fig:Hindi_results}
\end{figure}

\begin{figure}[htbp]
    \centering
    \includegraphics[width=\linewidth]{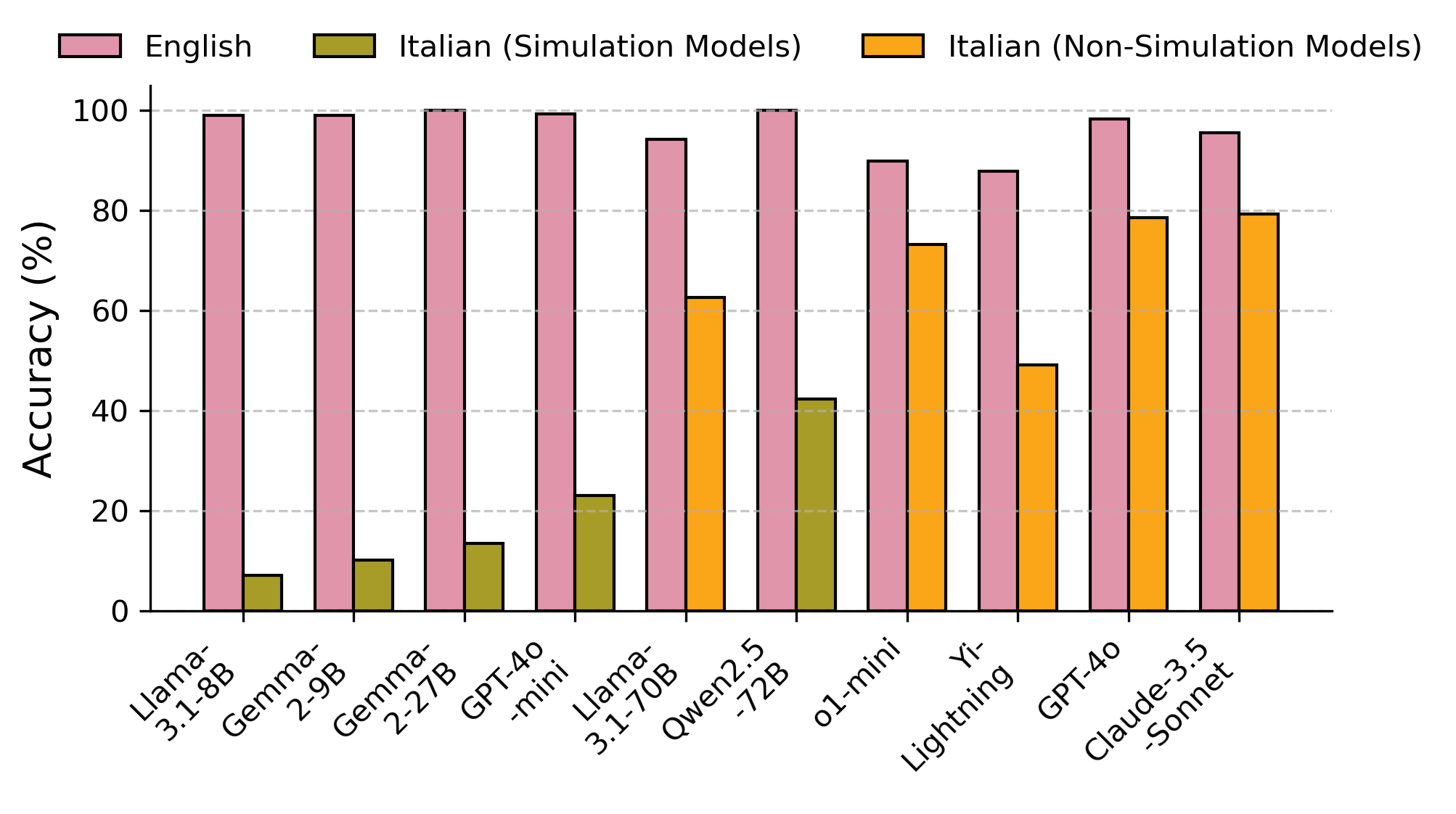}
    \caption{Performance of LLMs on English-Italian pairs in our candidate list.}
    \label{fig:Italian_results}
\end{figure}

\begin{figure}[htbp]
    \centering
    \includegraphics[width=\linewidth]{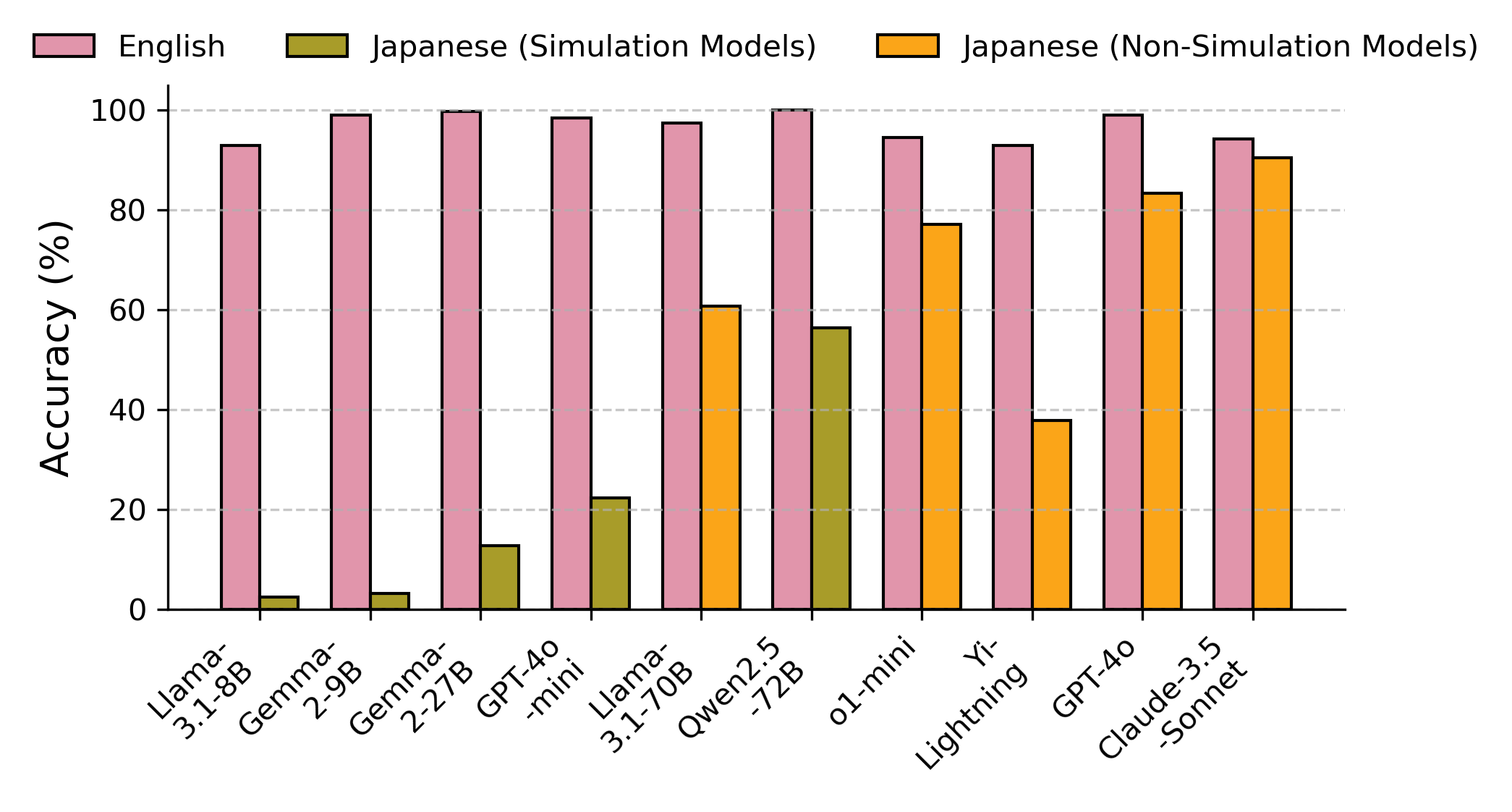}
    \caption{Performance of LLMs on English-Japanese pairs in our candidate list.}
    \label{fig:Japanese_results}
\end{figure}

\begin{figure}[htbp]
    \centering
    \includegraphics[width=\linewidth]{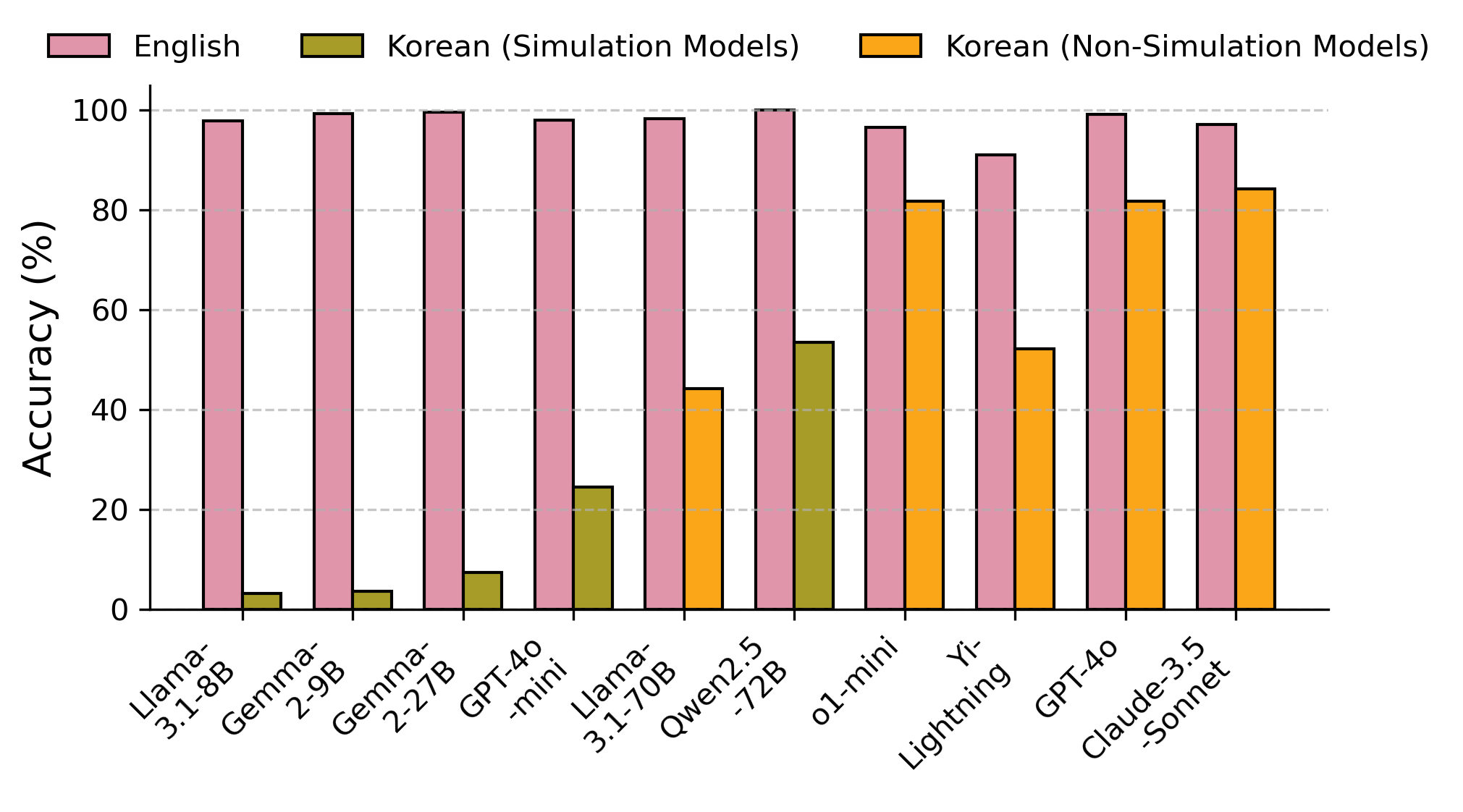}
    \caption{Performance of LLMs on English-Korean pairs in our candidate list.}
    \label{fig:Korean_results}
\end{figure}

\begin{figure}[htbp]
    \centering
    \includegraphics[width=\linewidth]{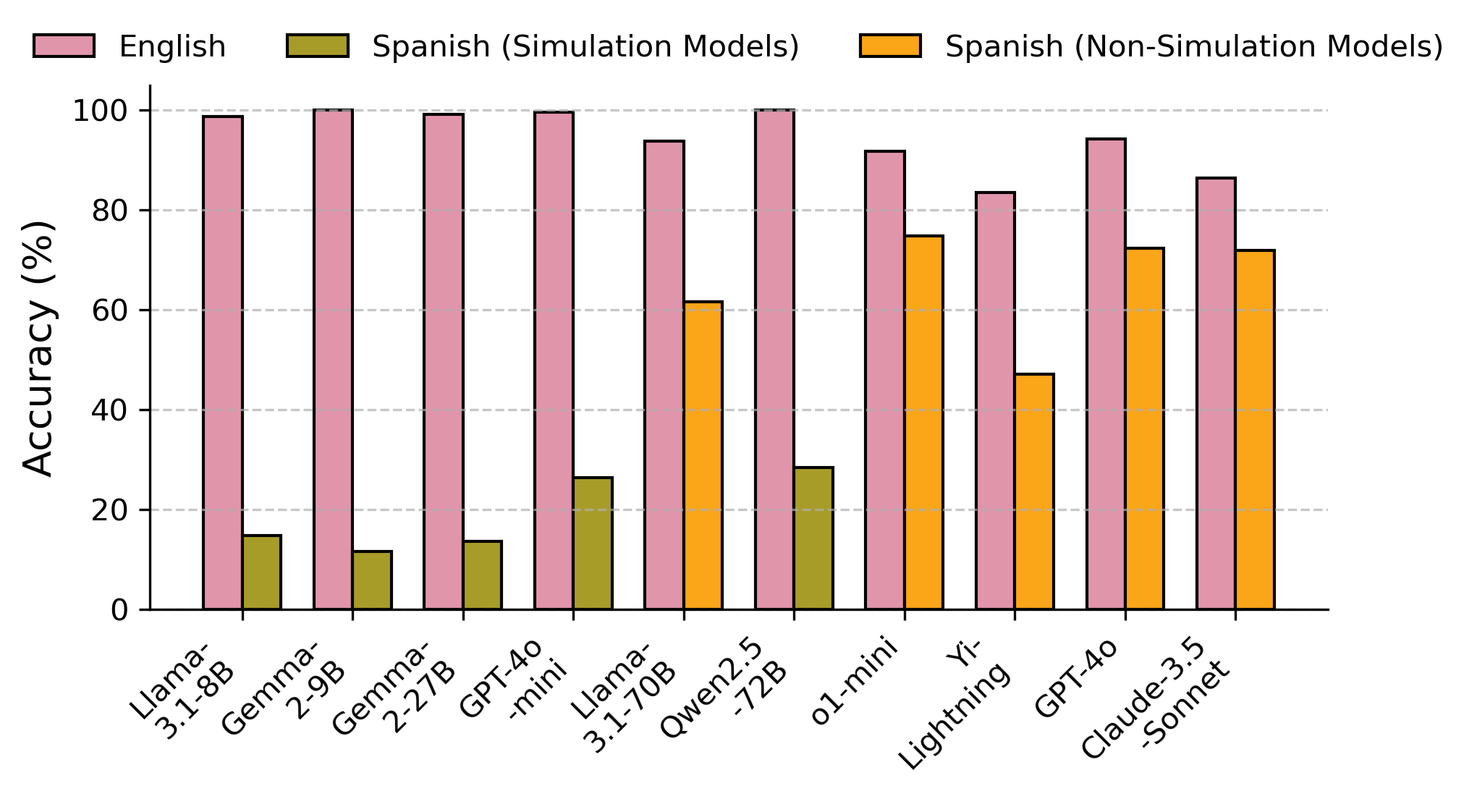}
    \caption{Performance of LLMs on English-Spanish pairs in our candidate list.}
    \label{fig:Spanish_results}
\end{figure}

\begin{figure}[htbp]
    \centering
    \includegraphics[width=\linewidth]{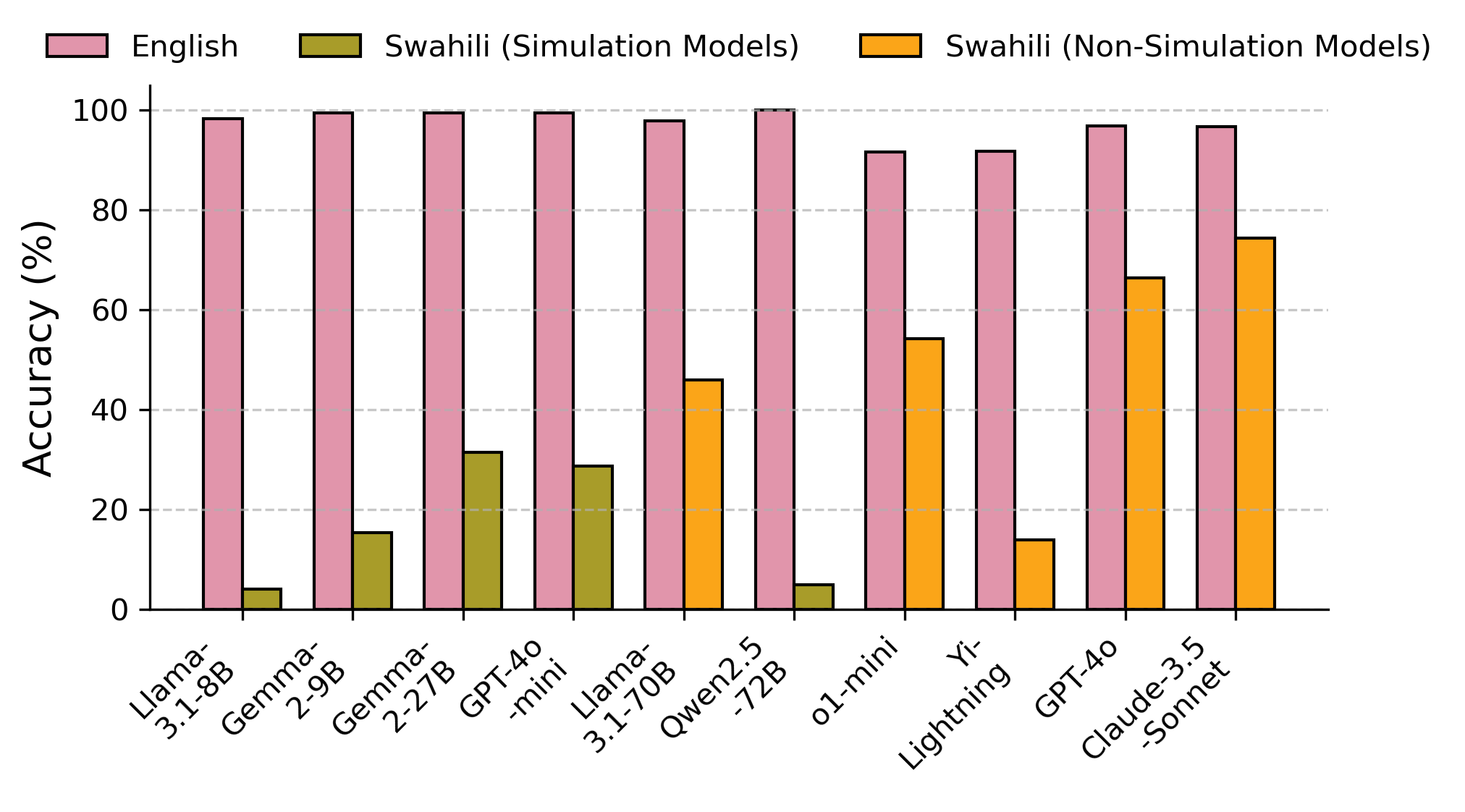}
    \caption{Performance of LLMs on English-Swahili pairs in our candidate list.}
    \label{fig:Swahili_results}
\end{figure}

\begin{figure}[htbp]
    \centering
    \includegraphics[width=\linewidth]{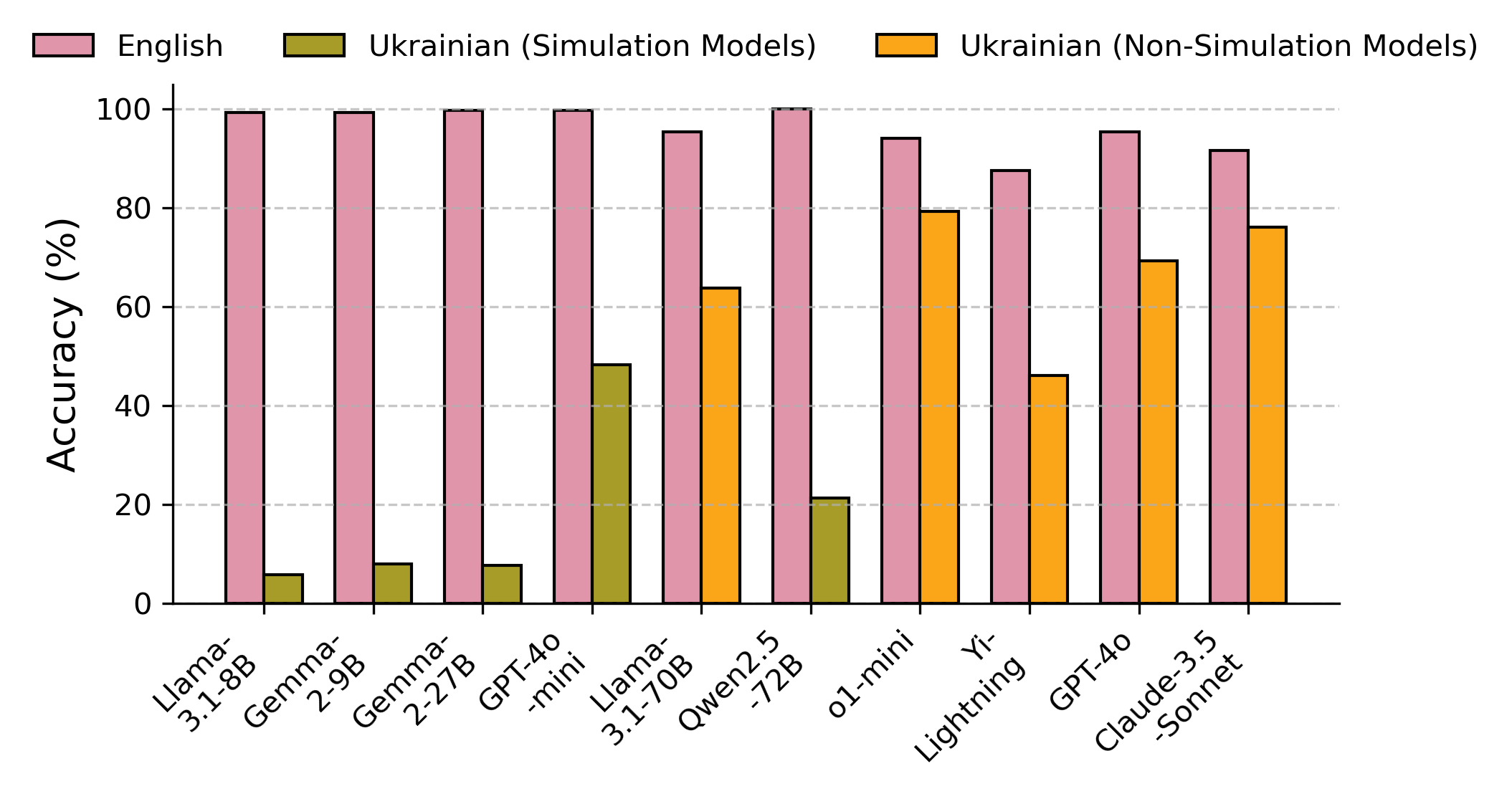}
    \caption{Performance of LLMs on English-Ukrainian pairs in our candidate list.}
    \label{fig:Ukrainian_results}
\end{figure}

\begin{figure}[htbp]
    \centering
    \includegraphics[width=\linewidth]{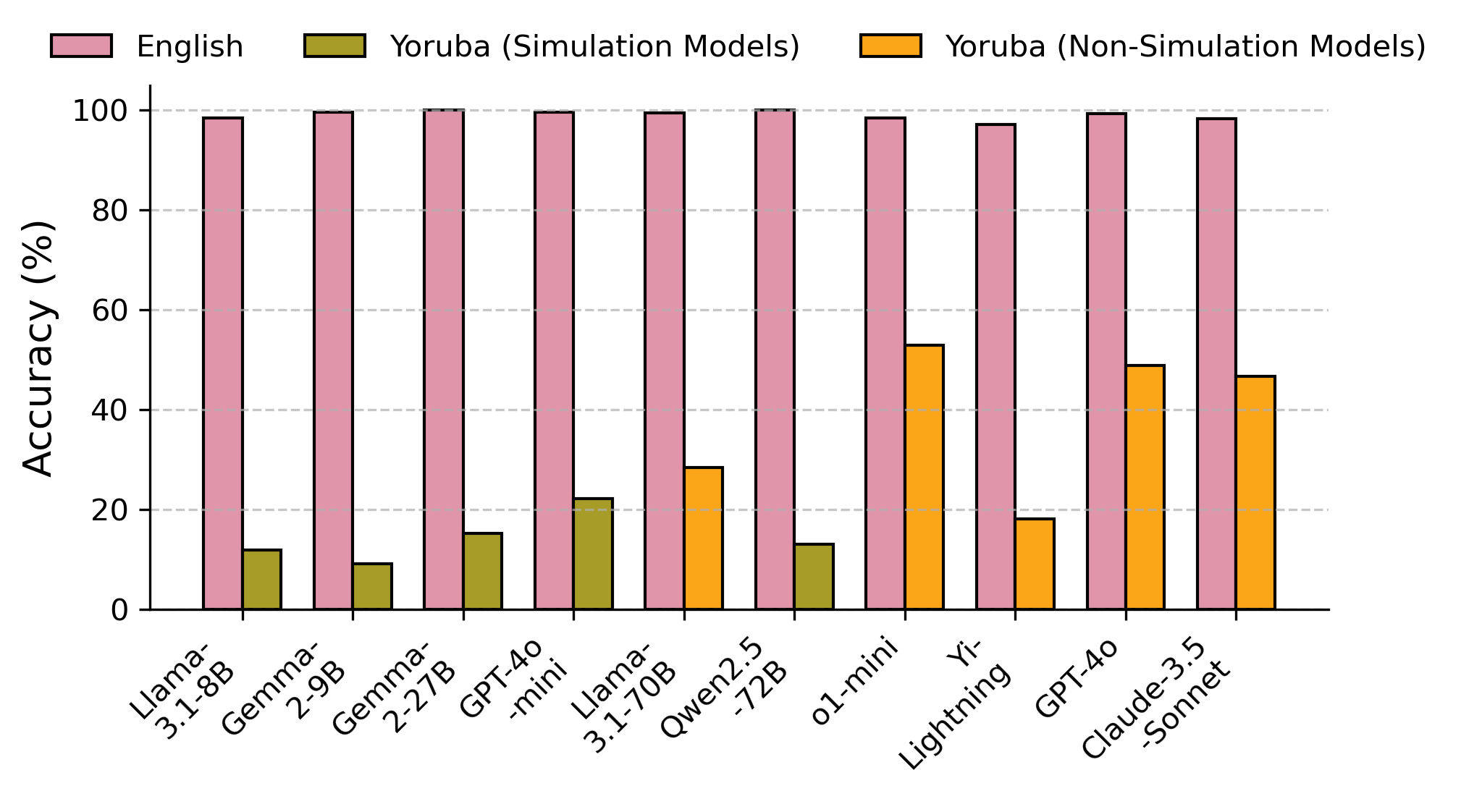}
    \caption{Performance of LLMs on English-Yoruba pairs in our candidate list.}
    \label{fig:Yoruba_results}
\end{figure}

\begin{figure}[htbp]
    \centering
    \includegraphics[width=\linewidth]{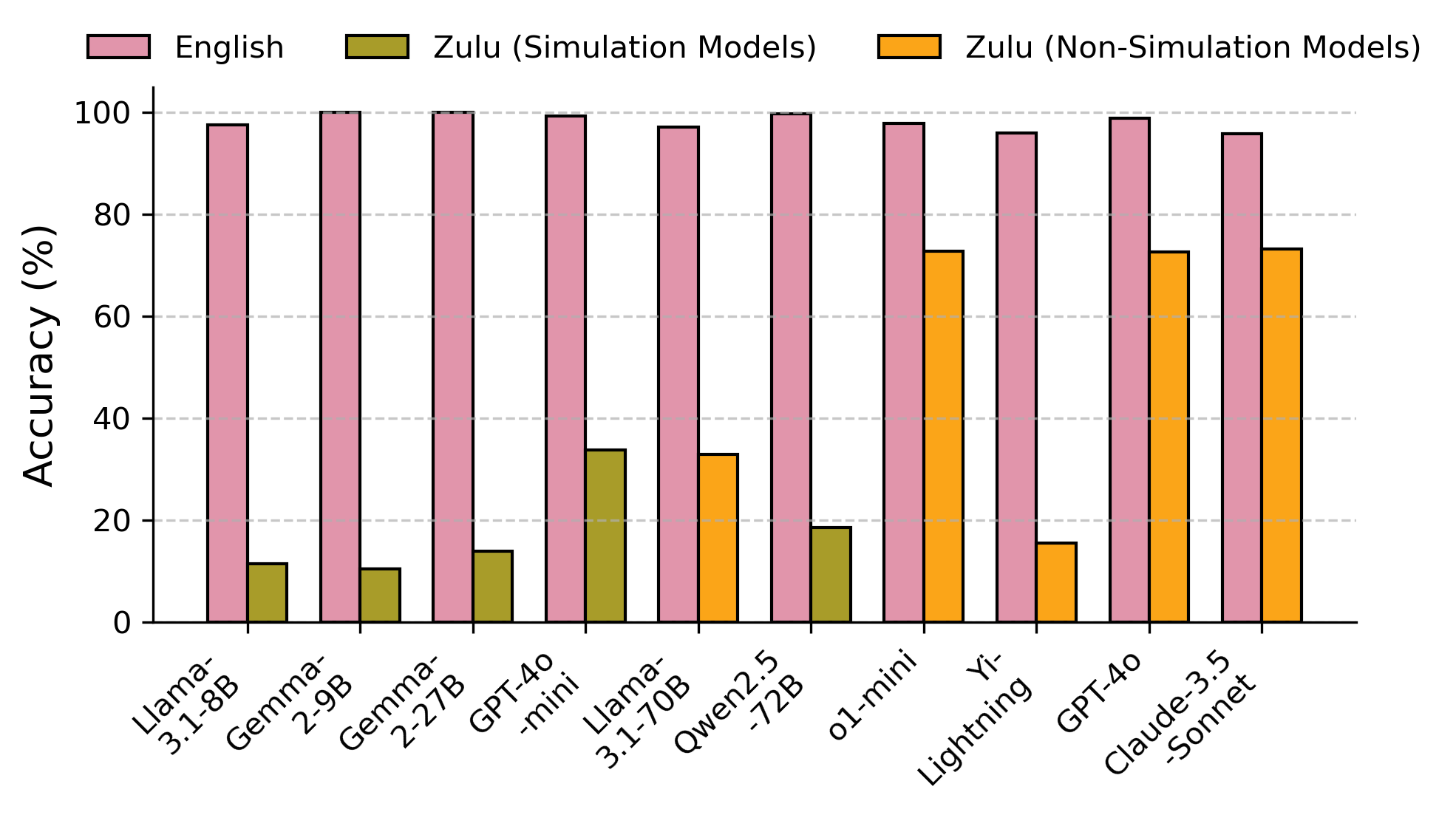}
    \caption{Performance of LLMs on English-Zulu pairs in our candidate list.}
    \label{fig:Zulu_results}
\end{figure}

\textbf{Experiment procedure of cross-Lingual weakness identification.} To generate bilingual question pairs for our cross-lingual weakness identification experiments, we employed LLM-based simulation using the following models: Llama-3.1-8B, Gemma-2-9B, Gemma-2-27B, GPT-4o-mini, and Qwen2.5-72B. This process resulted in a total of 6713 bilingual pairs across the following languages: Chinese (342 pairs), Japanese (314 pairs), Korean (456 pairs), French (312 pairs), Spanish (242 pairs), Italian (295 pairs), Ukrainian (323 pairs), German (322 pairs), Bengali (431 pairs), Hindi (327 pairs), Arabic (424 pairs), Hebrew (319 pairs), Amharic (665 pairs), Yoruba (813 pairs), Swahili (417 pairs), and Zulu (711 pairs). Subsequently, we performed zero-shot evaluations on all generated question pairs using the following models: Llama-3.1-8B, Gemma-2-9B, Gemma-2-27B, GPT-4o-mini, Llama-3.1-70B, Qwen2.5-72B, o1-mini, Yi-Lightning, GPT-4o, and Claude-3.5-sonnet. The results of these evaluations are presented in Figure \ref{fig:Chinese_results}, \ref{fig:Amharic_results}, \ref{fig:Arabic_results}, \ref{fig:Bengali_results},  \ref{fig:French_results}, \ref{fig:German_results}, \ref{fig:Hebrew_results}, \ref{fig:Hindi_results}, \ref{fig:Italian_results}, \ref{fig:Japanese_results}, \ref{fig:Korean_results}, \ref{fig:Spanish_results}, \ref{fig:Swahili_results}, \ref{fig:Ukrainian_results}, \ref{fig:Yoruba_results} and \ref{fig:Zulu_results} .

\textbf{Experiment procedure of quantifying the linguistic relationships.} To quantify the linguistic relationships between languages, we randomly sampled 100 generated bilingual pairs for each of the following languages: Chinese, Japanese, Korean, French, Spanish, Italian, Ukrainian, German, Bengali, Hindi, Arabic, Hebrew, Amharic, Yoruba, Swahili, and Zulu. We then translated the original question component of these pairs into each of the other fifteen languages using GPT-4o, and the perturbed question component using Google Translate's API \citep{googletranslateapi}. This process, along with the original language, resulted in a total of 25,600 bilingual pairs (16 languages * 100 pairs * 16 translations). We performed zero-shot evaluations on these pairs using six models: Llama-3.1-8B, Gemma-2-27B, GPT-4o-mini, Llama-3.1-70B, Qwen2.5-72B, and GPT-4o. The Relative Affinity Score was then calculated based on the average accuracy of these models, as shown in \autoref{fig:heatmap}.

\definecolor{headerbg}{RGB}{220, 230, 240}
\definecolor{rowgray}{RGB}{235, 240, 245} 
\definecolor{highlightorange}{RGB}{240, 150, 90}

\begin{table*}[t]
\centering
\small
\renewcommand{\arraystretch}{1.2}
\setlength{\tabcolsep}{3pt}
\caption{Percentage distribution of weaknesses across different categories for each language, compared to overall averages. Percentages exceeding the overall average for each category are highlighted in \textcolor{highlightorange}{orange}.  Column abbreviations are as follows: Sci \& Tech (Science \& Technology), Gen Knowl. (General Knowledge), Geo \& Env. (Geography \& Environment), Soc \& Cult. (Society \& Culture), Arts \& Lit. (Arts \& Literature), and Hist \& Events (History \& World Events).}
\label{tab:category}
\rowcolors{2}{rowwhite}{rowgray} 
\begin{tabular}{lcccccc}
    \toprule[1.5pt]
    \rowcolor{headerbg}
    \textcolor{black}{\textbf{Language}} & 
    \textcolor{black}{\textbf{Sci \& Tech}} &
    \textcolor{black}{\textbf{Gen Knowl.}} &
    \textcolor{black}{\textbf{Geo \& Env.}} &
    \textcolor{black}{\textbf{Soc \& Cult.}} &
    \textcolor{black}{\textbf{Arts \& Lit.}} &
    \textcolor{black}{\textbf{Hist \& Events}} \\
    \midrule[0.8pt]
    \midrule[0.8pt]
    Amharic   & \textcolor{highlightorange}{61.95\%} & 8.12\%   & 5.41\%   & 15.94\% & 2.26\%   & \textcolor{highlightorange}{6.32\%}   \\
    Arabic    & \textcolor{highlightorange}{55.42\%} & \textcolor{highlightorange}{15.57\%} & 0.71\%   & 9.20\%     & \textcolor{highlightorange}{9.91\%}   & \textcolor{highlightorange}{9.20\%}     \\
    Bengali   & \textcolor{highlightorange}{46.17\%} & \textcolor{highlightorange}{16.47\%} & \textcolor{highlightorange}{10.21\%} & 17.87\% & \textcolor{highlightorange}{7.66\%}   & 1.62\%   \\
    Chinese   & 25.73\% & 6.43\%   & \textcolor{highlightorange}{7.60\%}    & \textcolor{highlightorange}{47.08\%} & \textcolor{highlightorange}{12.28\%} & 0.88\%   \\
    French    & 37.50\%   & \textcolor{highlightorange}{13.78\%} & \textcolor{highlightorange}{10.26\%} & \textcolor{highlightorange}{26.28\%} & 6.73\%   & \textcolor{highlightorange}{5.45\%}   \\
    German    & 42.24\% & \textcolor{highlightorange}{19.88\%} & \textcolor{highlightorange}{9.32\%}   & 22.36\% & 0.62\%   & \textcolor{highlightorange}{5.59\%}   \\
    Hebrew    & 43.26\% & 10.02\%   & 0.58\%       & 22.57\% & \textcolor{highlightorange}{16.30\%}   & \textcolor{highlightorange}{6.27\%}   \\
    Hindi     & 44.95\% & \textcolor{highlightorange}{17.74\%} & 3.06\%   & 20.49\% & \textcolor{highlightorange}{11.31\%} & 2.45\%   \\
    Italian   & \textcolor{highlightorange}{49.15\%} & 6.10\%     & 3.39\%   & \textcolor{highlightorange}{26.10\%}   & 5.76\%   & \textcolor{highlightorange}{9.49\%}   \\
    Japanese  & \textcolor{highlightorange}{48.09\%} & \textcolor{highlightorange}{15.29\%} & \textcolor{highlightorange}{12.42\%} & 19.11\% & 4.46\%   & 0.64\%   \\
    Korean    & 27.41\% & \textcolor{highlightorange}{16.45\%} & \textcolor{highlightorange}{23.25\%} & \textcolor{highlightorange}{25.66\%} & 3.07\%   & 4.17\%   \\
    Spanish   & 23.97\% & \textcolor{highlightorange}{18.18\%} & 4.55\%   & 19.01\% & \textcolor{highlightorange}{21.90\%}     & \textcolor{highlightorange}{12.40\%}    \\
    Swahili   & \textcolor{highlightorange}{47.96\%} & 8.87\%   & 3.12\%   & \textcolor{highlightorange}{30.94\%} & 4.32\%   & 4.80\%    \\
    Ukrainian & 39.01\% & 10.53\% & 4.33\%   & \textcolor{highlightorange}{34.98\%} & 6.81\%   & 4.33\%   \\
    Yoruba    & \textcolor{highlightorange}{53.01\%} & 6.52\%   & \textcolor{highlightorange}{7.87\%}   & 21.03\% & 4.67\%   & \textcolor{highlightorange}{6.89\%}   \\
    Zulu      & \textcolor{highlightorange}{47.40\%}   & \textcolor{highlightorange}{13.36\%} & 0.98\%   & \textcolor{highlightorange}{25.60\%}   & \textcolor{highlightorange}{8.44\%}   & 4.22\%   \\
    \textbf{Overall Average} & 45.36\% & 12.20\% & 6.63\% & 23.40\% & 7.15\% & 5.26\% \\
    \bottomrule[1.5pt]
\end{tabular}
\end{table*}

\textbf{Experiment procedure of linguistic relationship analysis through fine-tuning.} Leveraging the English-Chinese and English-French question pairs generated in our dataset, we performed SFT and DPO on several Large Language Models: Llama-3.1-8B, Qwen2.5-7B, Gemma-2-9B, and Phi-3.5-Mini. For each model, we conducted separate fine-tuning runs using both the Chinese and French datasets. To ensure consistency across experiments, we trained for 4 epochs with a learning rate of 2.0e-4, employing a cosine learning rate scheduler and a warmup ratio of 0.1. The per-device training batch size was set to 1, with gradient accumulation performed over 8 steps. During training, we used the correct answers from the models' responses as the target output for each question. For evaluation, we used 10\% of the training data as a validation set, evaluated every 200 steps, and set the per-device evaluation batch size to 1.

\begin{table*}[htbp]
  \centering
  \small
  \renewcommand{\arraystretch}{1.1}
  \setlength{\tabcolsep}{6pt}
  \caption{Performance comparison of Gemma-2-9B and Qwen2.5-7B after SFT and DPO on French and Chinese datasets. The table shows evaluation results on various evaluation languages (EL), with the Asian language group highlighted in blue.  Performance differences (Diff.) are shown compared to the original model (Orig.).  "S Enh." represents the model enhanced by SFT, and "D Enh." represents the model enhanced by simulated DPO.}
  \label{tab:ft_full_dpo_supp}
  \begin{tabular}{lcccccccccc}
    \toprule
    & & & \multicolumn{4}{c}{\textbf{French Fine-Tuning}} & \multicolumn{4}{c}{\textbf{Chinese Fine-Tuning}} \\
    \cmidrule(lr){4-7} \cmidrule(lr){8-11}
    \multirow{-2}{*}{\textbf{Model}} & \multirow{-2}{*}{\textbf{EL}} & \multirow{-2}{*}{\textbf{Orig.}} & \textbf{S Enh.} & \textbf{D Enh.} & \textbf{S Diff.} & \textbf{D Diff.} & \textbf{S Enh.} & \textbf{D Enh.} & \textbf{S Diff.} & \textbf{D Diff.} \\
    \midrule
    \multirow[c]{6}{*}{Gemma-2-9B} & French   & 0.222    & --    & --    & --    & --    & 0.510 & 0.522 & 0.288 & 0.300 \\
     & German   & 0.115 & 0.495 & 0.505 & 0.381 & 0.391 & 0.463 & 0.475 & 0.348 & 0.360 \\
     & Spanish  & 0.109 & 0.541 & 0.555 & 0.433 & 0.447 & 0.463 & 0.458 & 0.314 & 0.309 \\
     & \cellcolor{lightblue}Chinese & \cellcolor{lightblue}0.111 & \cellcolor{lightblue}0.552 & \cellcolor{lightblue}0.560 & \cellcolor{lightblue}0.441 & \cellcolor{lightblue}0.449 & \cellcolor{lightblue}--    & \cellcolor{lightblue}--    & \cellcolor{lightblue}--   & \cellcolor{lightblue}--    \\
     & \cellcolor{lightblue}Japanese& \cellcolor{lightblue}0.099 & \cellcolor{lightblue}0.527 & \cellcolor{lightblue}0.535 & \cellcolor{lightblue}0.428 & \cellcolor{lightblue}0.436 & \cellcolor{lightblue}0.576 & \cellcolor{lightblue}0.588 & \cellcolor{lightblue}0.478 & \cellcolor{lightblue}0.490 \\
     & \cellcolor{lightblue}Korean  & \cellcolor{lightblue}0.083 & \cellcolor{lightblue}0.421 & \cellcolor{lightblue}0.430 & \cellcolor{lightblue}0.338 & \cellcolor{lightblue}0.347 & \cellcolor{lightblue}0.537 & \cellcolor{lightblue}0.545 & \cellcolor{lightblue}0.454 & \cellcolor{lightblue}0.462 \\
    \midrule
    \multirow[c]{6}{*}{Qwen2.5-7B} & French   & 0.321    & --    & --    & --    & --    & 0.494 & 0.503 & 0.173 & 0.182 \\
     & German   & 0.233 & 0.447 & 0.455 & 0.214 & 0.222 & 0.491 & 0.485 & 0.258 & 0.252 \\
     & Spanish  & 0.145 & 0.537 & 0.548 & 0.393 & 0.403 & 0.426 & 0.432 & 0.281 & 0.287 \\
     & \cellcolor{lightblue}Chinese & \cellcolor{lightblue}0.281 & \cellcolor{lightblue}0.584 & \cellcolor{lightblue}0.595 & \cellcolor{lightblue}0.304 & \cellcolor{lightblue}0.314 & \cellcolor{lightblue}--    & \cellcolor{lightblue}--    & \cellcolor{lightblue}--   & \cellcolor{lightblue}--    \\
     & \cellcolor{lightblue}Japanese& \cellcolor{lightblue}0.194 & \cellcolor{lightblue}0.408 & \cellcolor{lightblue}0.415 & \cellcolor{lightblue}0.213 & \cellcolor{lightblue}0.221 & \cellcolor{lightblue}0.592 & \cellcolor{lightblue}0.600 & \cellcolor{lightblue}0.398 & \cellcolor{lightblue}0.406 \\
     & \cellcolor{lightblue}Korean  & \cellcolor{lightblue}0.140 & \cellcolor{lightblue}0.329 & \cellcolor{lightblue}0.337 & \cellcolor{lightblue}0.189 & \cellcolor{lightblue}0.197 & \cellcolor{lightblue}0.384 & \cellcolor{lightblue}0.393 & \cellcolor{lightblue}0.243 & \cellcolor{lightblue}0.252 \\
    \bottomrule
  \end{tabular}
\end{table*}

\section{Human Evaluation}
\label{appendix:human}
To ensure that the target language questions in our generated bilingual pairs maintained semantic equivalence and answer consistency with the original English questions, we conducted a human evaluation study. We randomly sampled 100 bilingual pairs from the candidate list for each of the following sixteen languages: Chinese, Japanese, Korean, French, Spanish, Italian, Ukrainian, German, Bengali, Hindi, Arabic, Hebrew, Amharic, Yoruba, Swahili, and Zulu. Four undergraduate students majoring in computer science, proficient in English and various translation tools, were divided into two groups to assess: (1) whether the target language question maintained semantic equivalence with the original English question, and (2) whether the answer to the target language question was consistent with the answer to the original English question. The results of this evaluation are summarized in \autoref{tab:human_evaluation}.

\begin{figure}[htbp]
    \centering
    \includegraphics[width=\linewidth]{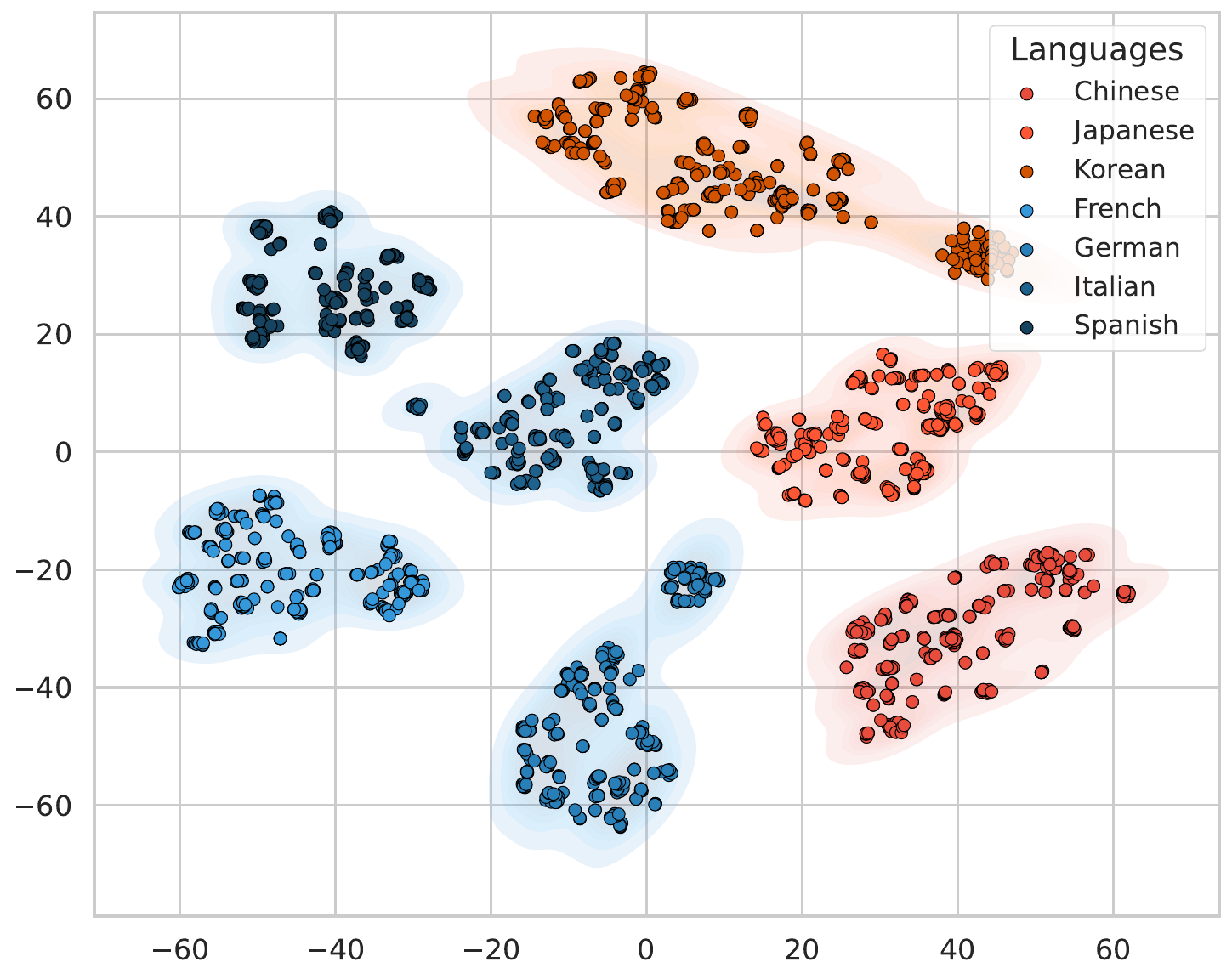}
    \caption{T-SNE visualization of the embeddings generated by LLaMA-3.1-8B for seven English–target language question pairs.}
    \label{fig:embedding_tsne}
\end{figure}



\section{Case Study}
\label{appendix:case}
\begin{figure*}[ht]
    \centering
    \includegraphics[width=0.7\linewidth]{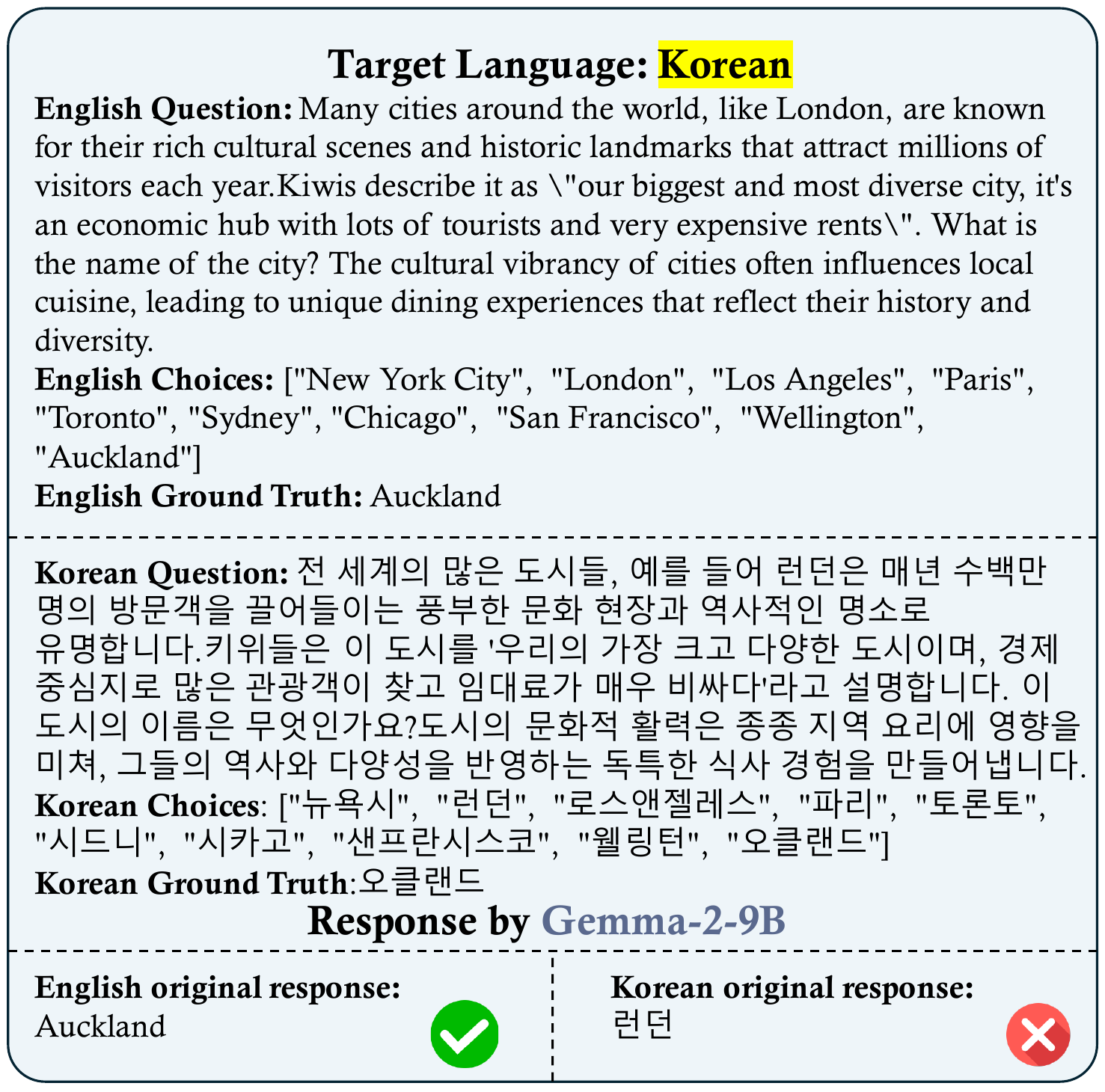}
    \caption{Case study: Gemma-2-9B's responses to English-Korean question pairs.}
    \label{fig:Korean_case}
\end{figure*}

\begin{figure*}[ht]
    \centering
    \includegraphics[width=0.7\linewidth]{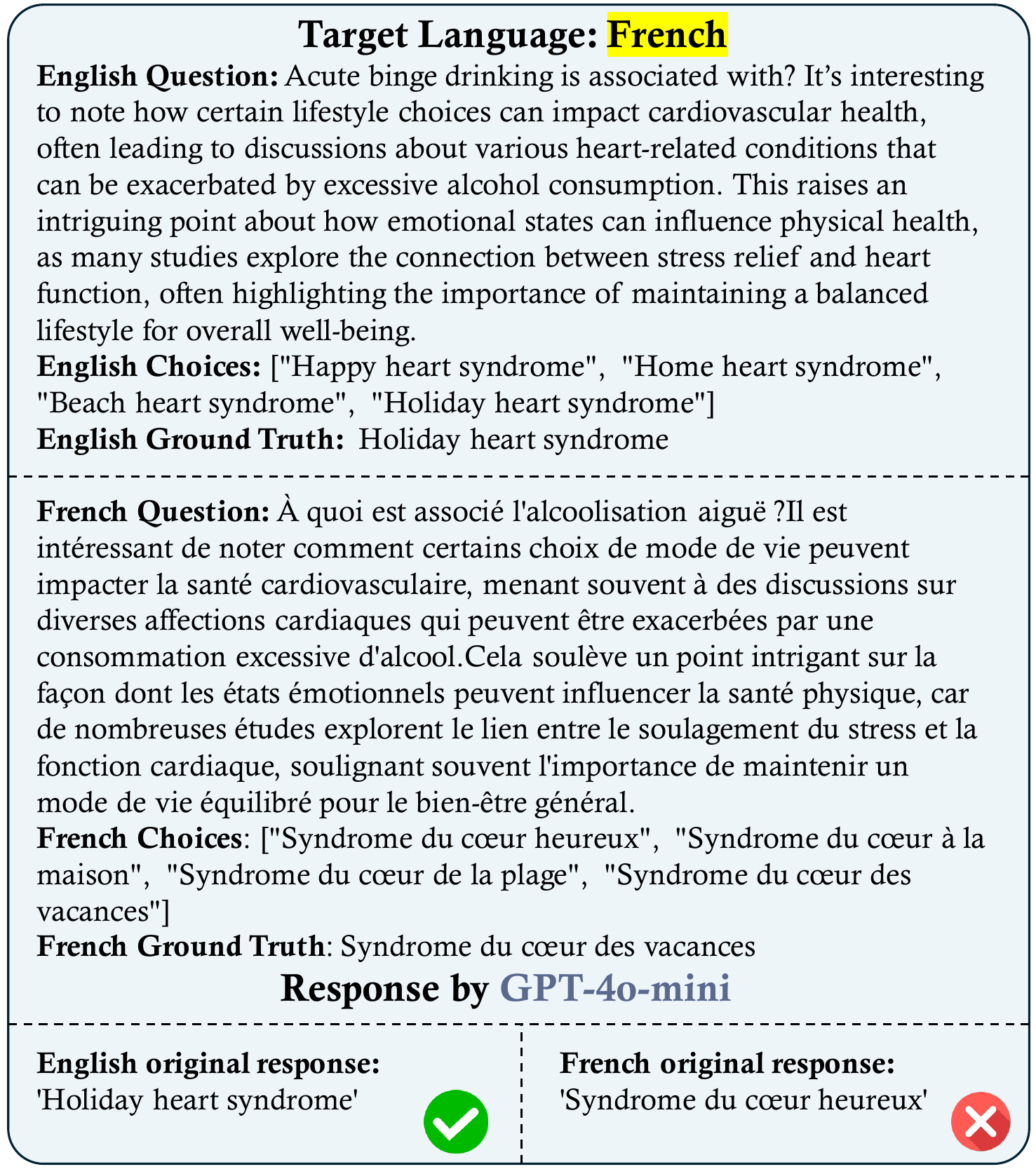}
    \caption{Case study: GPT-4o-mini's responses to English-French question pairs.}
    \label{fig:French_case}
\end{figure*}

\begin{figure*}[ht]
    \centering
    \includegraphics[width=0.7\linewidth]{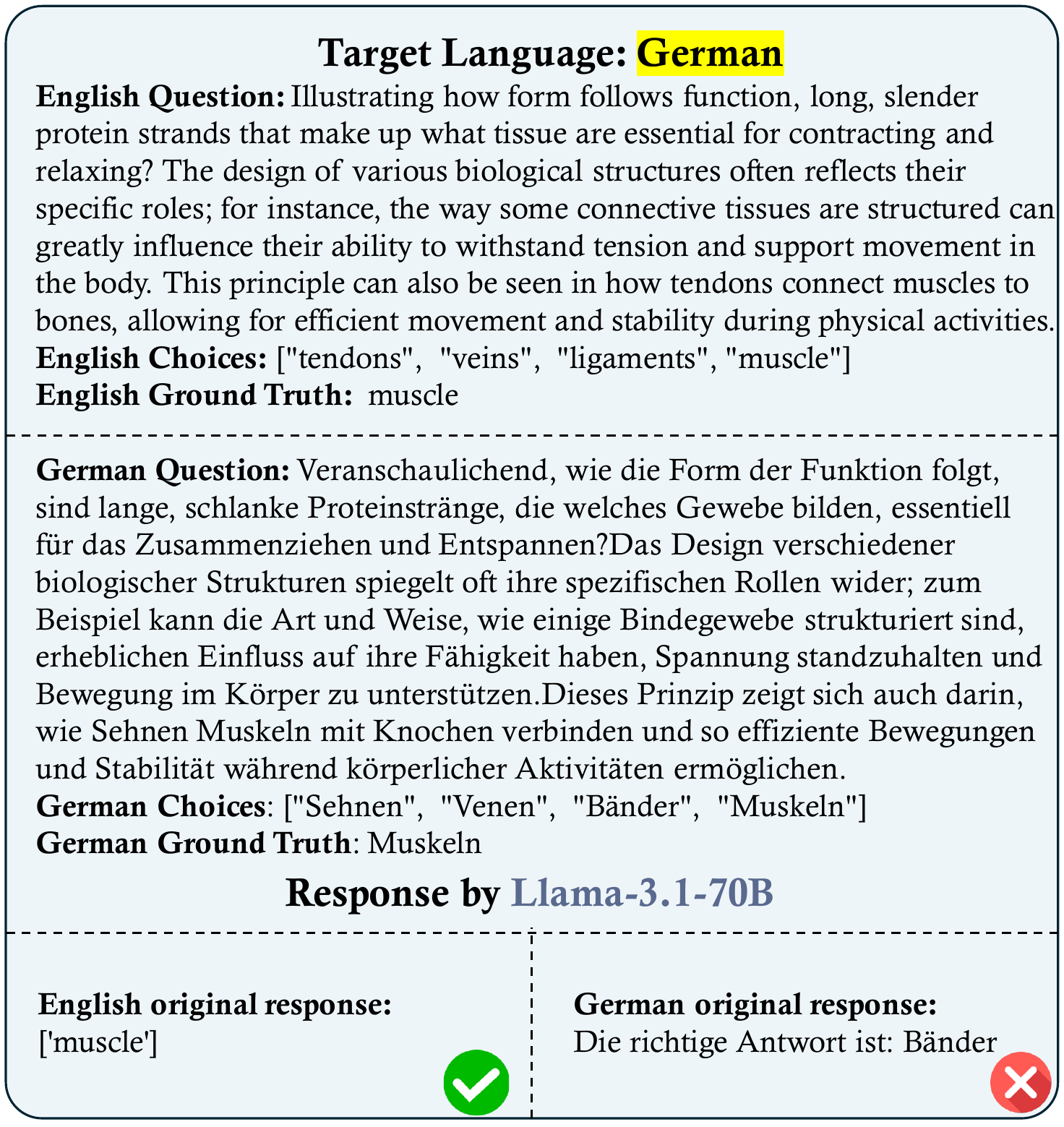}
    \caption{Case study: Llama-3.1-70B's responses to English-German question pairs.}
    \label{fig:German_case}
\end{figure*}

\begin{figure*}[ht]
    \centering
    \includegraphics[width=0.7\linewidth]{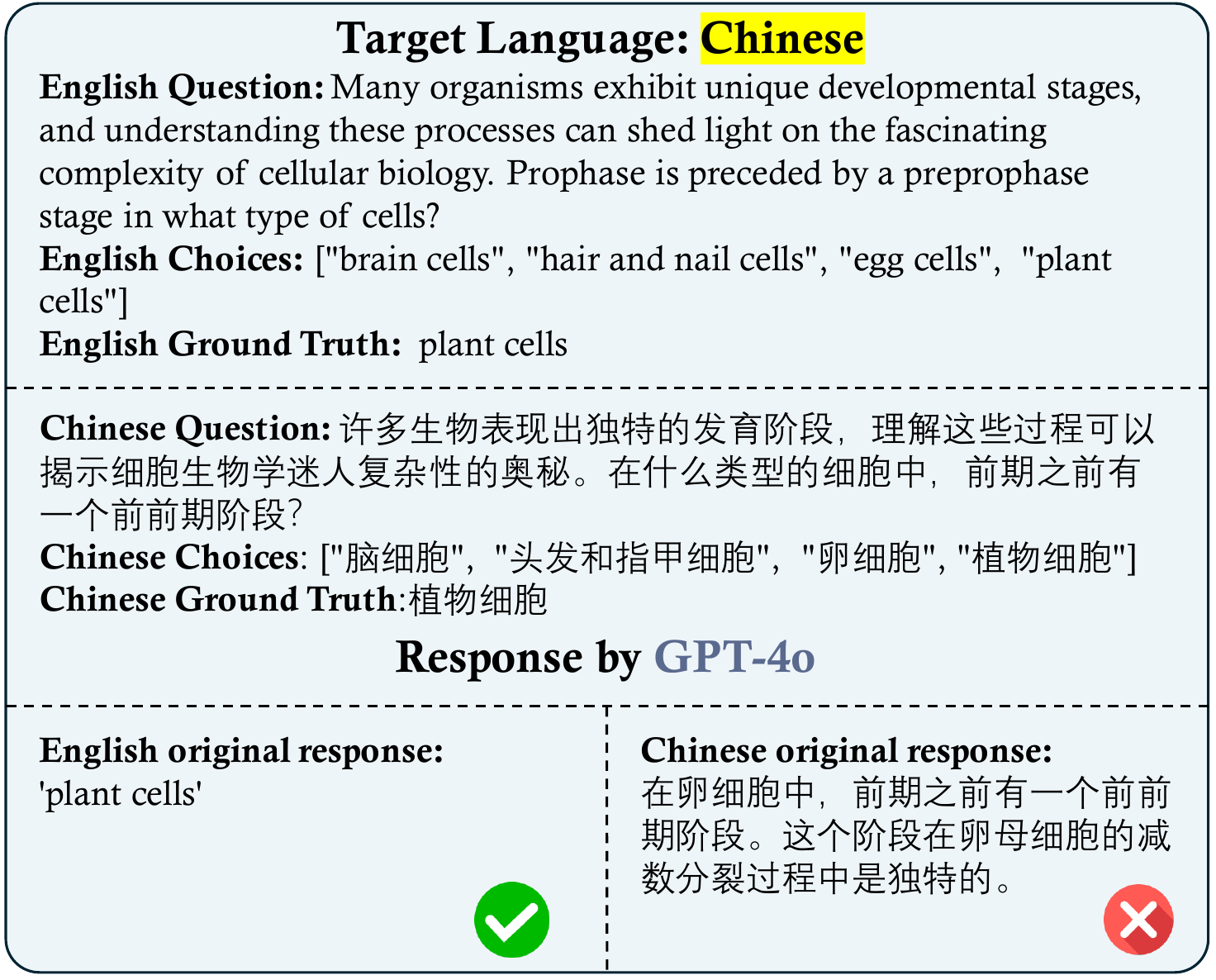}
    \caption{Case study: GPT-4o's responses to English-Chinese question pairs.}
    \label{fig:Chinese_case}
\end{figure*}

\begin{figure*}[ht]
    \centering
    \includegraphics[width=0.7\linewidth]{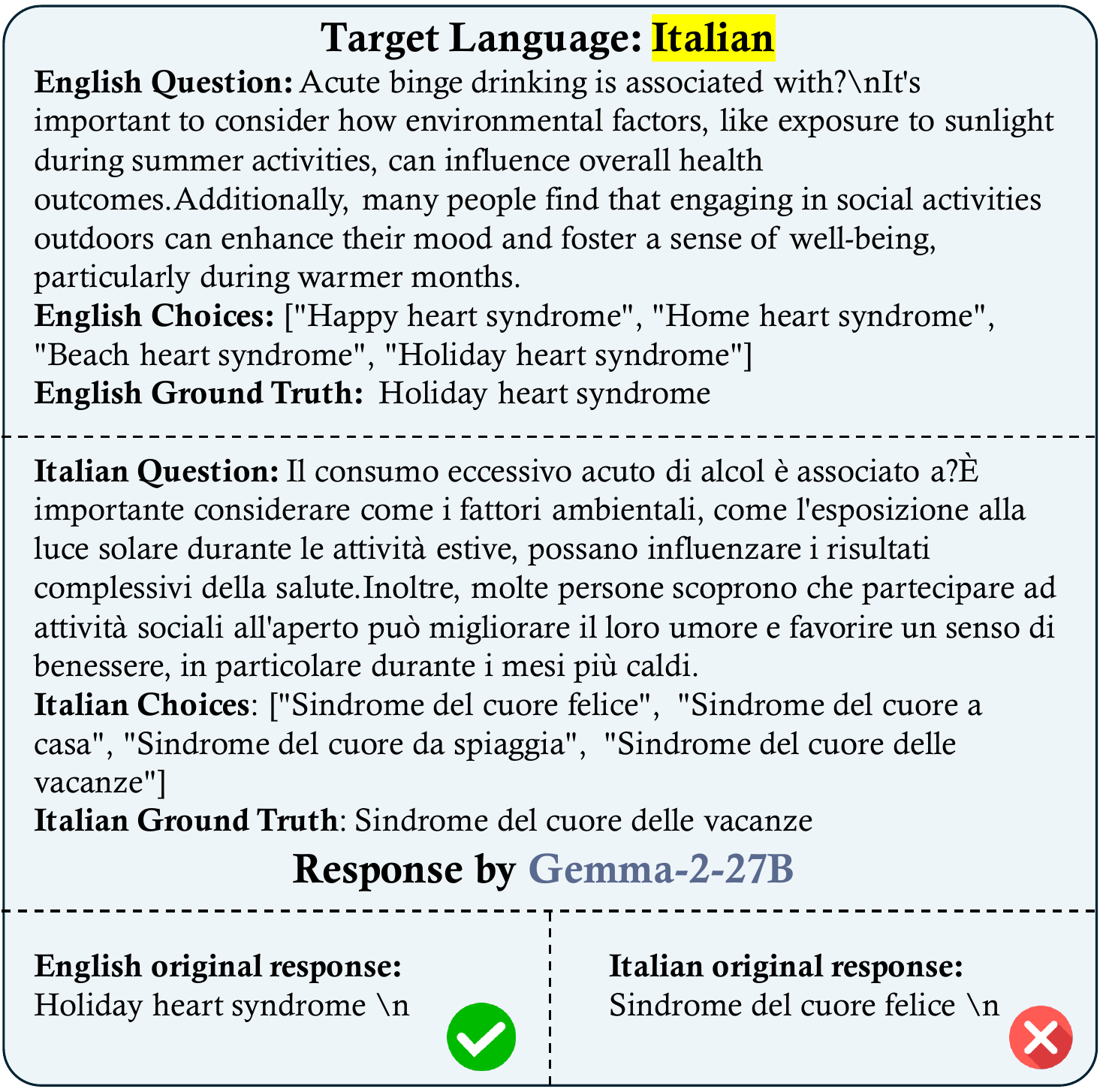}
    \caption{Case study: Gemma-2-27B's responses to English-Italian question pairs.}
    \label{fig:Italian_case}
\end{figure*}

\begin{figure*}[ht]
    \centering
    \includegraphics[width=0.7\linewidth]{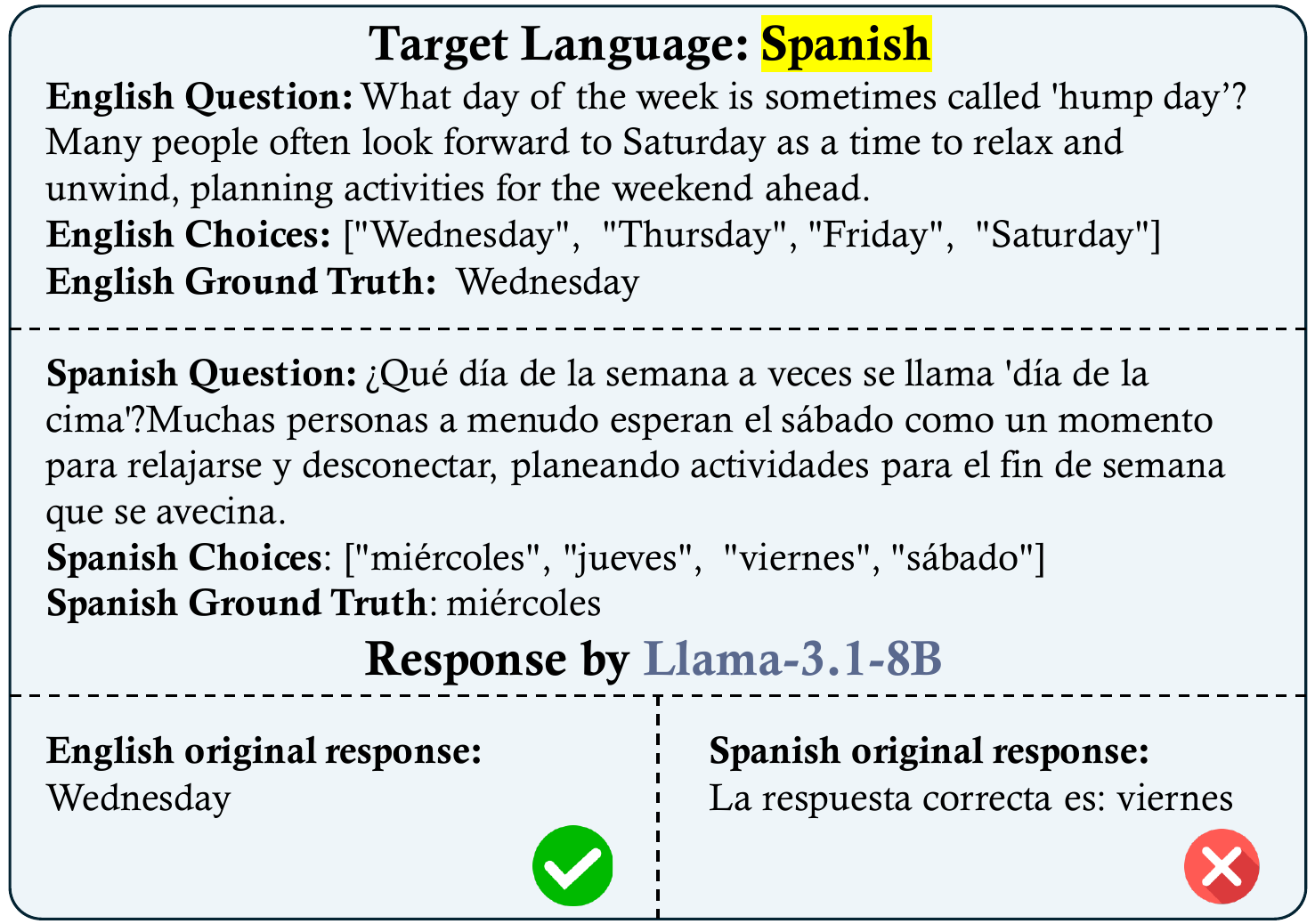}
    \caption{Case study: Llama-3.1-8B's responses to English-Spanish question pairs.}
    \label{fig:Spanish_case}
\end{figure*}

\begin{figure*}[ht]
    \centering
    \includegraphics[width=0.7\linewidth]{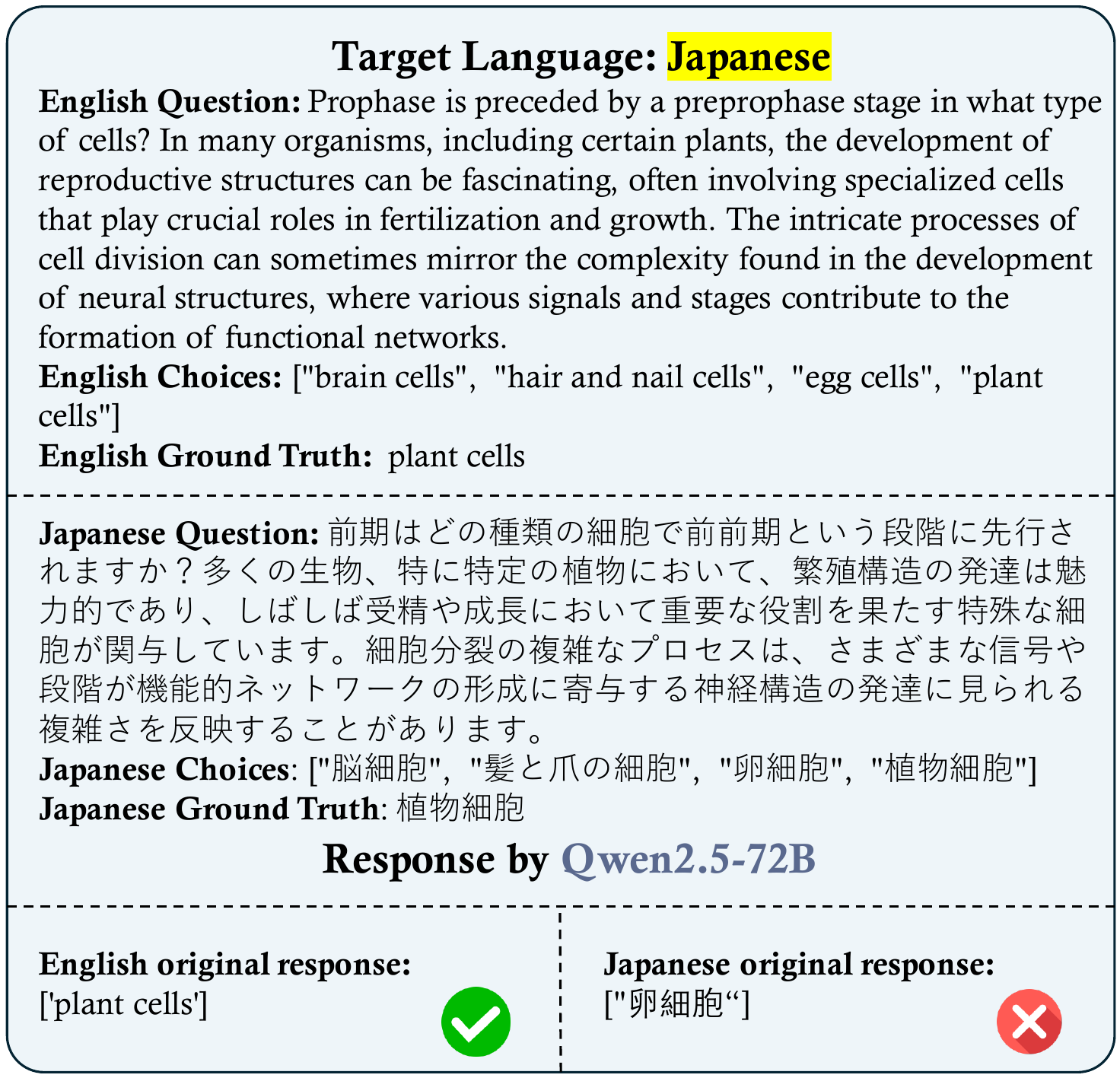}
    \caption{Case study: Qwen2.5-72B's responses to English-Japanese question pairs.}
    \label{fig:Japanese_case}
\end{figure*}

\begin{figure*}[ht]
    \centering
    \includegraphics[width=0.7\linewidth]{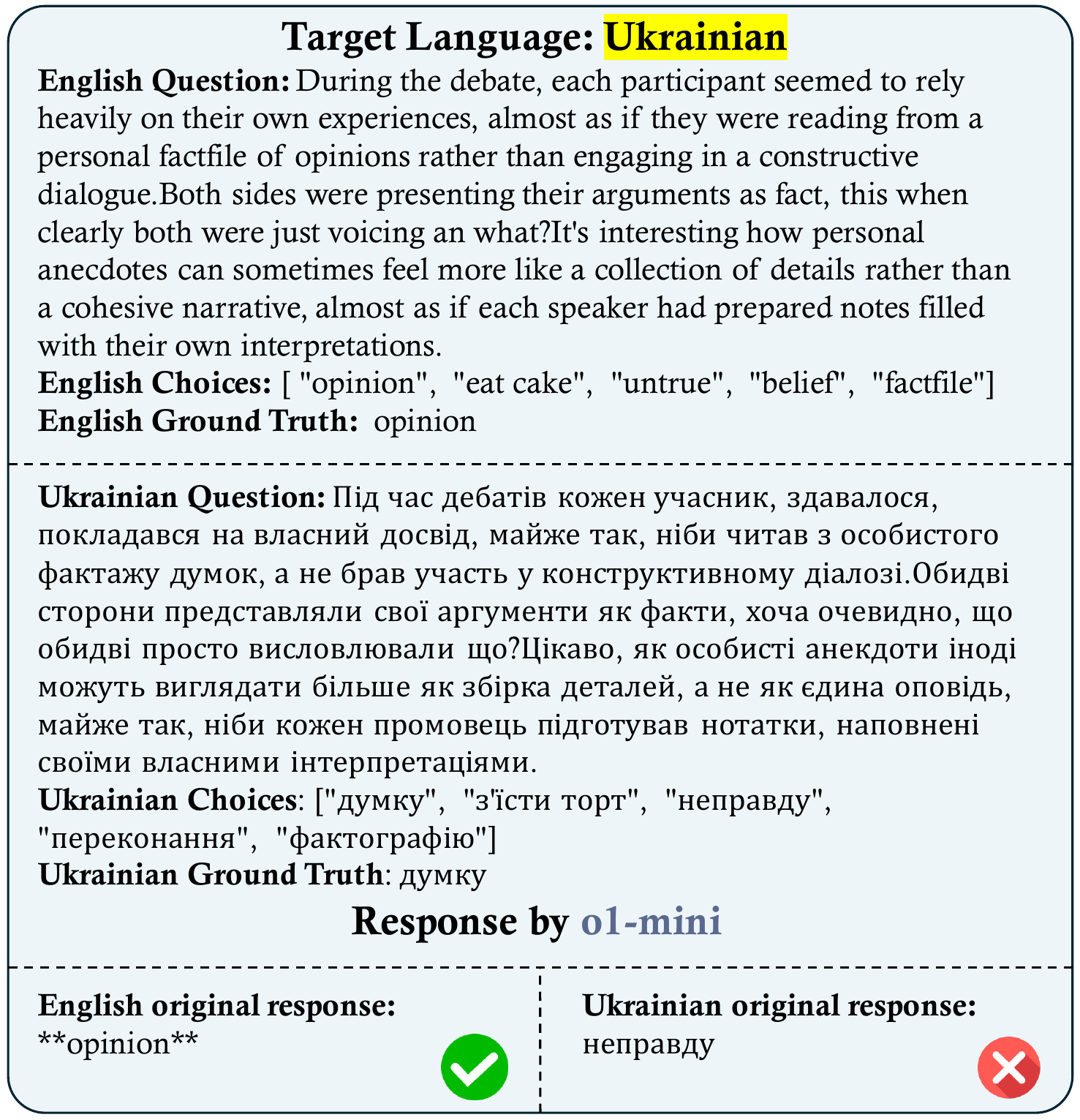} 
    \caption{Case study: o1-mini's responses to English-Ukrainian question pairs.}
    \label{fig:Ukrainian_case}
\end{figure*}

In Figure~\ref{fig:Korean_case}, \ref{fig:French_case}, \ref{fig:German_case}, \ref{fig:Chinese_case}, \ref{fig:Italian_case}, \ref{fig:Spanish_case}, \ref{fig:Japanese_case}, and \ref{fig:Ukrainian_case}, we illustrate case studies of model responses to English-target language (Korean, French, German, Chinese, Italian, Spanish, Japanese, and Ukrainian, respectively) question pairs.

\begin{table}[htbp]
    \centering
    \small
    \setlength{\tabcolsep}{4pt} 
    \renewcommand{\arraystretch}{1.1}
    \caption{Results of human evaluation on semantic equivalence (Semantic Eq.) and answer consistency (Answer Consis.) between original English questions and target language questions in bilingual pairs.}
    \label{tab:human_evaluation}
    \scalebox{0.9}{
    \begin{tabular}{p{2.5cm}cc} 
    \toprule[1pt]
    \textbf{Language} & \textbf{Semantic Eq. (\%)} & \textbf{Answer Consis. (\%)} \\
    \midrule
    \textbf{Amharic}          & 83.0 & 88.0 \\
    \textbf{Arabic}           & 90.0 & 94.0 \\
    \textbf{Bengali}          & 88.0 & 93.0 \\
    \textbf{Chinese}          & 95.0 & 98.0 \\
    \textbf{French}           & 97.0 & 99.0 \\
    \textbf{German}           & 96.0 & 98.0 \\
    \textbf{Hebrew}           & 93.0 & 95.0 \\
    \textbf{Hindi}            & 91.0 & 94.0 \\
    \textbf{Italian}          & 96.0 & 97.0 \\
    \textbf{Japanese}         & 93.0 & 95.0 \\
    \textbf{Korean}           & 91.0 & 93.0 \\
    \textbf{Spanish}          & 98.0 & 100.0 \\
    \textbf{Swahili}          & 89.0 & 93.0 \\
    \textbf{Ukrainian}        & 92.0 & 95.0 \\
    \textbf{Yoruba}           & 84.0 & 90.0 \\
    \textbf{Zulu}             & 86.0 & 91.0 \\
    \bottomrule[1pt]
    \end{tabular}}
\end{table}

\clearpage

\clearpage

\begin{figure}[htbp]
    \centering
    \includegraphics[width=\linewidth]{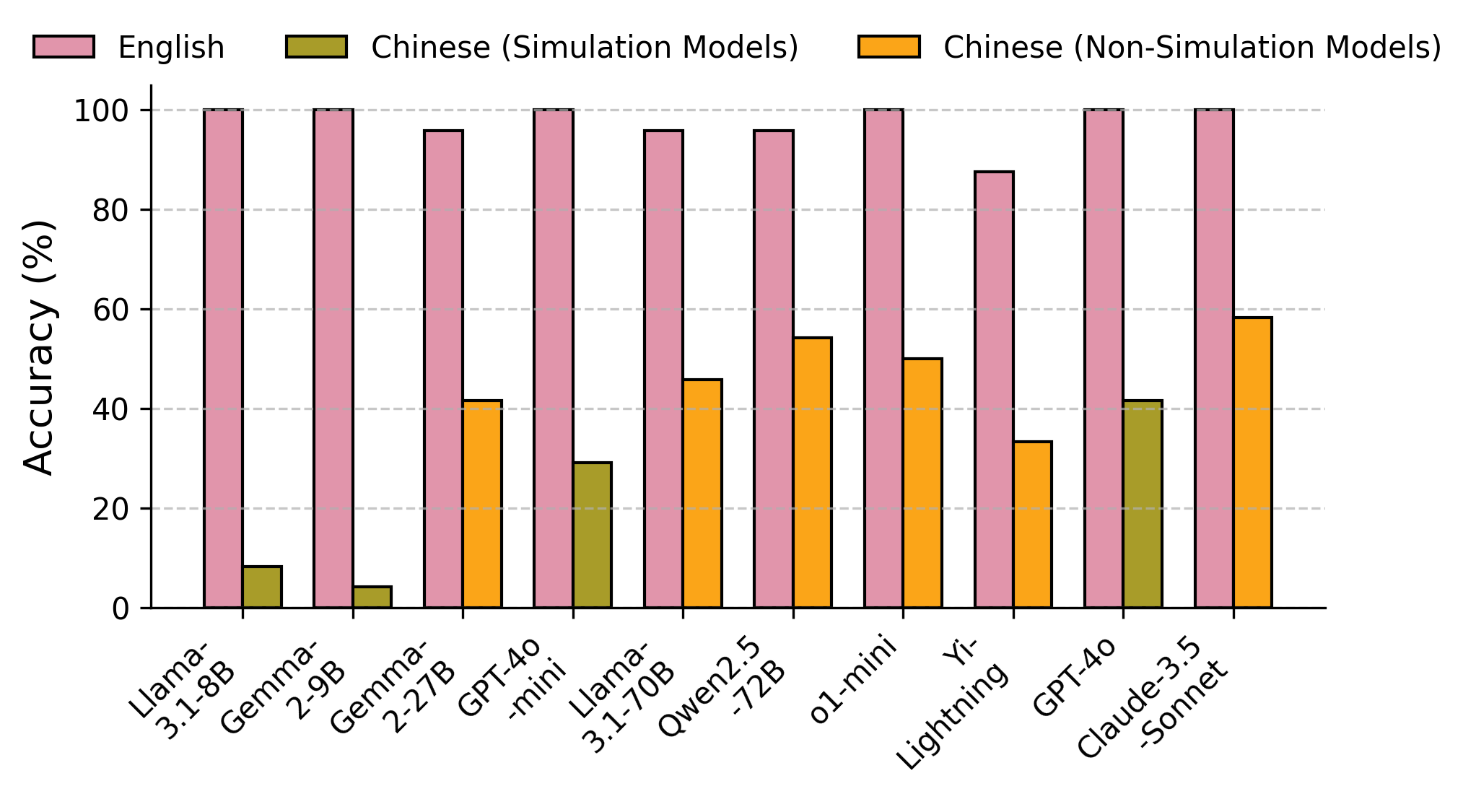}
    \caption{Performance of LLMs on English-Chinese pairs after incorporating GPT-4o into the LLM-Based simulation. The Chinese accuracy of GPT-4o dropped significantly}
    \label{fig:4o}
\end{figure}

\clearpage
\clearpage

\section{Prompt Template}
\label{appendix:prompt_template}

\begin{promptbox*}{Template for Generating Perturbation}
[Instruction] \newline
You are an expert at subtly embedding distractions based on the incorrect option provided. Your task is to generate a distraction that aligns with the incorrect option without altering the original question's quality or meaning. Follow these specific rules:  \newline

1. The distraction should naturally integrate with the context of the question but must not explicitly introduce incorrect information or contradict the correct answer.  
2. The distraction must be subtle and should not make it obvious that it is related to the incorrect option.  \newline

[The Start of the Question] \newline
\{question\} \newline
[The End of the Question] \newline

[The Start of the Model's Answer] \newline
\{answer\} \newline
[The End of the Model's Answer] \newline

[The Start of the Incorrect Option] \newline
\{wrong\_answer\} \newline
[The End of the Incorrect Option] \newline

[Output Format] \newline
\{Generated Distraction: \textless Provide a subtle, contextually relevant distraction based on the incorrect option \textgreater\}
\end{promptbox*}







\clearpage  
\begin{promptbox*}{Template for English-to-French Translation}  
[Instruction] \newline  
Vous êtes un traducteur professionnel. Votre tâche consiste à traduire le texte, les choix et la réponse ci-dessous de manière précise et naturelle en français, tout en conservant le sens original des questions et des choix. Veuillez respecter strictement les règles suivantes :  \newline  

- La traduction des réponses et des choix doit refléter fidèlement le sens original, sans aucune altération, omission ou ajout.  \newline  
- Toutes les phrases comportant un point d'interrogation doivent rester sous forme de question après traduction, sans changer le ton ou la structure de la phrase.  \newline  
- Le contenu traduit doit respecter les normes et usages de la langue française, être fluide et naturel, en évitant les traductions littérales ou maladroites.  \newline  

[The Start of the Text] \newline  
\{question\} \newline  
[The End of the Text] \newline  

[The Start of the Choices] \newline  
\{choices\} \newline  
[The End of the Choices] \newline  

[The Start of the Answer] \newline  
\{ground\_truth\} \newline  
[The End of the Answer] \newline  

[Output Format] \newline  
\{"text": "\textless Texte traduit en français\textgreater", "choices": ["\textless Choix traduit en français 1\textgreater", "\textless Choix traduit en français 2\textgreater", ...], "answer": "\textless Réponse traduite en français\textgreater"\}  
\end{promptbox*}

\clearpage  
\begin{promptbox*}{Template for English-to-German Translation}  
[Instruction] \newline  
Sie sind ein professioneller Übersetzungsexperte. Ihre Aufgabe besteht darin, den folgenden Text, die Auswahlmöglichkeiten und die Antwort präzise und natürlich ins Deutsche zu übersetzen, wobei der ursprüngliche Sinn der Frage und der Auswahlmöglichkeiten erhalten bleiben muss. Halten Sie sich strikt an die folgenden Regeln:  \newline  

- Die Übersetzung der Antworten und Auswahlmöglichkeiten muss den ursprünglichen Sinn vollständig bewahren, ohne jegliche Abweichungen, Hinzufügungen oder Kürzungen.  \newline  
- Alle Sätze mit einem Fragezeichen müssen auch nach der Übersetzung die Form einer Frage beibehalten, ohne den Ton oder die Struktur des Satzes zu verändern.  \newline  
- Der übersetzte Inhalt muss den sprachlichen Gepflogenheiten des Deutschen entsprechen, natürlich und flüssig formuliert sein und wörtliche, ungeschmeidige Übersetzungen vermeiden.  \newline  

[The Start of the Text] \newline  
\{question\} \newline  
[The End of the Text] \newline  

[The Start of the Choices] \newline  
\{choices\} \newline  
[The End of the Choices] \newline  

[The Start of the Answer] \newline  
\{ground\_truth\} \newline  
[The End of the Answer] \newline  

[Output Format] \newline  
\{"text": "\textless Übersetzter Text\textgreater", "choices": ["\textless Übersetzte Auswahl1\textgreater", "\textless Übersetzte Auswahl2\textgreater", ...], "answer": "\textless Übersetzte Antwort\textgreater"\}  
\end{promptbox*}

\clearpage  
\begin{promptbox*}{Template for English-to-Italian Translation}  
[Instruction] \newline  
Sei un traduttore professionista. Il tuo compito è tradurre il seguente testo, le opzioni e la risposta in italiano in modo accurato e naturale, assicurandoti di preservare il significato originale della domanda e delle opzioni. Segui rigorosamente le seguenti regole:  \newline  

- La traduzione delle risposte e delle opzioni deve mantenere completamente il significato originale, senza alcuna deviazione, aggiunta o omissione.  \newline  
- Tutte le frasi con un punto interrogativo devono mantenere la forma interrogativa dopo la traduzione, senza alterare il tono o la struttura della frase.  \newline  
- Il contenuto tradotto deve rispettare le abitudini linguistiche dell'italiano, essere naturale e fluido, evitando traduzioni letterali e rigide.  \newline  

[The Start of the Text] \newline  
\{question\} \newline  
[The End of the Text] \newline  

[The Start of the Choices] \newline  
\{choices\} \newline  
[The End of the Choices] \newline  

[The Start of the Answer] \newline  
\{ground\_truth\} \newline  
[The End of the Answer] \newline  

[Output Format] \newline  
\{"text": "\textless Testo tradotto\textgreater", "choices": ["\textless Opzione tradotta 1\textgreater", "\textless Opzione tradotta 2\textgreater", ...], "answer": "\textless Risposta tradotta\textgreater"\}  
\end{promptbox*}

\clearpage  
\begin{promptbox*}{Template for English-to-Spanish Translation}  
[Instruction] \newline  
Eres un experto en traducción profesional. Tu tarea es traducir el siguiente texto, opciones y respuestas de manera precisa y natural al español, asegurándote de conservar el significado original de las preguntas y opciones. Por favor, cumple estrictamente con las siguientes reglas:  \newline  

- La traducción de las respuestas y opciones debe conservar completamente el significado original, sin desviaciones ni adiciones.  \newline  
- Todas las oraciones que contengan un signo de interrogación deben mantener la forma de pregunta en la traducción, sin cambiar el tono ni la estructura de la oración.  \newline  
- El contenido traducido debe ajustarse a las costumbres del idioma español, expresándose de manera natural y fluida, evitando traducciones literales.  \newline  

[The Start of the Text] \newline  
\{question\} \newline  
[The End of the Text] \newline  

[The Start of the Choices] \newline  
\{choices\} \newline  
[The End of the Choices] \newline  

[The Start of the Answer] \newline  
\{ground\_truth\} \newline  
[The End of the Answer] \newline  

[Output Format] \newline  
\{"text": "\textless texto traducido\textgreater", "choices": ["\textless opción traducida 1\textgreater", "\textless opción traducida 2\textgreater", ...], "answer": "\textless respuesta traducida\textgreater"\}  
\end{promptbox*}

\begin{promptbox*}{Template for Answering English Questions (Zero-Shot + CoT)}  
[Instruction] \newline  
Please carefully read the question below and provide a solution from the choices. You must choose the model's final answer from one of the choices. Let's think step by step!  \newline  

[The Start of the Question] \newline  
\{question\} \newline  
[The End of the Question] \newline  

[The Start of the Choices] \newline  
\{choices\} \newline  
[The End of the Choices] \newline  

[Output Format] \newline  
\{"final\_answer": "\textless Your selected answer, exactly matching one of the given choices\textgreater"\}  
\end{promptbox*}





\clearpage  
\begin{promptbox*}{Template for Answering French Questions (Zero-Shot + CoT)}  
[Instruction] \newline  
Veuillez lire attentivement la question ci-dessous et choisir une réponse parmi les options proposées. Votre réponse finale doit correspondre exactement à l'une des options données. Réfléchissons étape par étape ! \newline  

[Début de la question] \newline  
\{question\} \newline  
[Fin de la question] \newline  

[Début des options] \newline  
\{choices\} \newline  
[Fin des options] \newline  

[Format de sortie] \newline  
\{"final\_answer": "\textless Votre réponse finale, correspondant exactement à l'une des options données\textgreater"\}  
\end{promptbox*}

\begin{promptbox*}{Template for Answering Italian Questions (Zero-Shot + CoT)}  
[Instruction] \newline  
Leggi attentamente la domanda qui sotto e fornisci una soluzione scegliendo tra le opzioni disponibili. La tua risposta finale deve corrispondere esattamente a una delle opzioni fornite. Pensiamo passo dopo passo!  \newline  

[Inizio della Domanda] \newline  
\{question\} \newline  
[Fine della Domanda] \newline  

[Inizio delle Opzioni] \newline  
\{choices\} \newline  
[Fine delle Opzioni] \newline  

[Formato di Output] \newline  
\{"final\_answer": "\textless La tua risposta finale, che deve corrispondere esattamente a una delle opzioni date\textgreater"\}  
\end{promptbox*}  

\begin{promptbox*}{Template for Answering Spanish Questions (Zero-Shot + CoT)}  
[Instruction] \newline  
Por favor, lee atentamente la siguiente pregunta y proporciona una solución eligiendo una de las opciones dadas. Tu respuesta final debe coincidir exactamente con una de las opciones. ¡Pensemos paso a paso!  \newline  

[Inicio de la Pregunta] \newline  
\{question\} \newline  
[Fin de la Pregunta] \newline  

[Inicio de las Opciones] \newline  
\{choices\} \newline  
[Fin de las Opciones] \newline  

[Formato de Salida] \newline  
\{"final\_answer": "\textless Tu respuesta final, exactamente igual a una de las opciones dadas\textgreater"\}  
\end{promptbox*}

\clearpage
\begin{promptbox*}{Template for Extracting Answer}
[Instruction] \newline
You are an expert in answer selecting. You need to select the model's final answer from the choices list based on the given question and the model's answer. \newline

[The Start of the Question] \newline
\{question\} \newline
[The End of the Question] \newline

[The Start of the Model's Answer] \newline
\{answer\} \newline
[The End of the Model's Answer] \newline

[The Start of the Choices] \newline
\{choices\} \newline
[The End of the Choices] \newline

[Output Format] \newline
\{"final\_answer": \textless Your extracted answer, strictly the same as the option in choices\textgreater\}
\end{promptbox*}

\begin{promptbox*}{Template for Adding Direct Perturbation.}
[Instruction] \newline
You are perturbation design expert. Add contextually relevant but non-essential information related to the topic in the question. The added content must NOT affect the problem's answerability or the validity of choices. Maintain original question structure verbatim. \newline

[The Start of the Question] \newline
\{question\} \newline
[The End of the Question] \newline

[Requirements] \newline
1. Add 2-3 background sentences before the original question. \newline
2. Include 1-2 practical application examples after the question. \newline
3. Keep all technical terms but expand their explanations. \newline
4. Preserve original question wording. \newline
5. NEVER mention or include any answer choices. \newline
6. Omit any reference to multiple-choice options. \newline

[Output Format] \newline
New question: \textless Your modified question WITHOUT ANY CHOICES\textgreater
\end{promptbox*}

\end{document}